\documentclass{article}

\usepackage[final,nonatbib]{neurips_2022}

\usepackage[utf8]{inputenc} % allow utf-8 input
\usepackage[T1]{fontenc}    % use 8-bit T1 fonts
\usepackage[colorlinks]{hyperref}       % hyperlinks
\usepackage{url}            % simple URL typesetting
\usepackage{booktabs}       % professional-quality tables
\usepackage{amsfonts}       % blackboard math symbols
\usepackage{nicefrac}       % compact symbols for 1/2, etc.
\usepackage{microtype}      % microtypography
\newcommand{\ie}{\textit{i}.\textit{e}.}
\usepackage{amsmath}
\usepackage{multirow}
\usepackage[table,xcdraw]{xcolor}
\usepackage{bm}
\usepackage{paralist} 
\usepackage{graphicx}
\usepackage{wrapfig}
\usepackage{bbding}
\usepackage{multicol}
\usepackage{color}
\usepackage{amssymb}

\title{Intermediate Prototype Mining Transformer for Few-Shot Semantic Segmentation}

\author{
 Yuanwei Liu$^{1}$\footnotemark[1]
 \hspace{15pt}
 Nian Liu$^{2}$\footnotemark[1]
 \hspace{15pt}
 Xiwen Yao$^{1}$
 \hspace{15pt}
 Junwei Han$^{1}$\footnotemark[2]
 \hspace{15pt}
 \\
 $^1$Northwestern Polytechnical University
  \\
 $^2$Mohamed bin Zayed University of Artificial Intelligence
 \\
 {\tt\small
    \{liuyuanwei9809, liunian228, yaoxiwen517, junweihan2010\}@gmail.com
    }\\
}

\begin{document}

\maketitle
\footnotetext[1]{Equal contribution.} %对应脚注[1]
\footnotetext[2]{Corresponding author.} %对应脚注[2]

\begin{abstract}
Few-shot semantic segmentation aims to segment the target objects in query under the condition of a few annotated support images. Most previous works strive to mine more effective category information from the support to match with the corresponding objects in query. However, they all ignored the category information gap between query and support images. If the objects in them show large intra-class diversity, forcibly migrating the category information from the support to the query is ineffective. To solve this problem, we are the first to introduce an intermediate prototype for mining both deterministic category information from the support and adaptive category knowledge from the query. Specifically, we design an Intermediate Prototype Mining Transformer (IPMT) to learn the prototype in an iterative way. In each IPMT layer, we propagate the object information in both support and query features to the prototype and then use it to activate the query feature map. By conducting this process iteratively, both the intermediate prototype and the query feature can be progressively improved. At last, the final query feature 
is used to yield precise segmentation prediction. Extensive experiments on both PASCAL-${5^{i}} $ and COCO-$ 20^{i} $ datasets clearly verify the effectiveness of our IPMT and show that it outperforms previous state-of-the-art methods by a large margin. Code is available at \href{https://github.com/LIUYUANWEI98/IPMT}{https://github.com/LIUYUANWEI98/IPMT}
\end{abstract}

\vspace{-1.5mm}
\section{Introduction}
\label{intro}
\vspace{-1.5mm}

% Owing to the rapid development of the deep learning techniques, the field of computer vision has witnessed great progress in recent years. However, these achievements rely heavily on a large amount of annotated data, while collecting such data is time-consuming and labor-intensive work.
Recent great progress on computer vision rely heavily on a large amount of annotated data, the collecting of which is a time-consuming and labor-intensive work.
% Meanwhile, in some way, the quality of data annotation closely influences the performance of the neural network, and this phenomenon is more obvious in label-intensive dense prediction tasks, such as semantic segmentation and instance segmentation.
% Therefore, it is necessary to seek a task that does not rely on amount of labeled data. As a paradigm 
To solve this problem, few-shot learning is proposed to learn a model that can be generalized to novel categories with only a few annotated images. This setting is also closer to human learning habits which can learn knowledge from scarce annotated examples and identify new categories quickly.

In this paper, we focus on the few-shot semantic segmentation (FSS) task which aims to segment novel objects in the query image with a few annotated support samples. 
% In order to better generalize on novel categories under the condition of a few samples, FSS follows the meta-learning framework which has success in few-shot classification. Specifically, in this framework, the model is trained on seen category with amount of labels to mimic the testing on unseen category with a few labels. 
Currently, a lot of works have been proposed for FSS and many of them are based on prototype learning.
% they can be divided into three types, \ie, prototype-based, graph-based, and transformer-based. 
These methods \cite{zhang2019canet,tian2020prior,li2021adaptive} extract prototypes from the support set to represent the category information and then match them with the query features in a matching network to perform segmentation.
% Graph-based methods \cite{zhang2019pyramid,wang2020few,xie2021scale} construct graph networks to pass message between query features which is  and support images. Transformer-based methods \cite{lu2021simpler,zhang2021few} utilize transformer to aggregate pixel-wise support features into queries.
Other graph-based methods \cite{zhang2019pyramid,wang2020few,xie2021scale} and transformer-based methods \cite{lu2021simpler,zhang2021few} share the similar high-level idea to convey the category information from the support set to the query image.

\begin{figure*}[ht]
	\begin{center}
		\includegraphics[scale = 0.3]{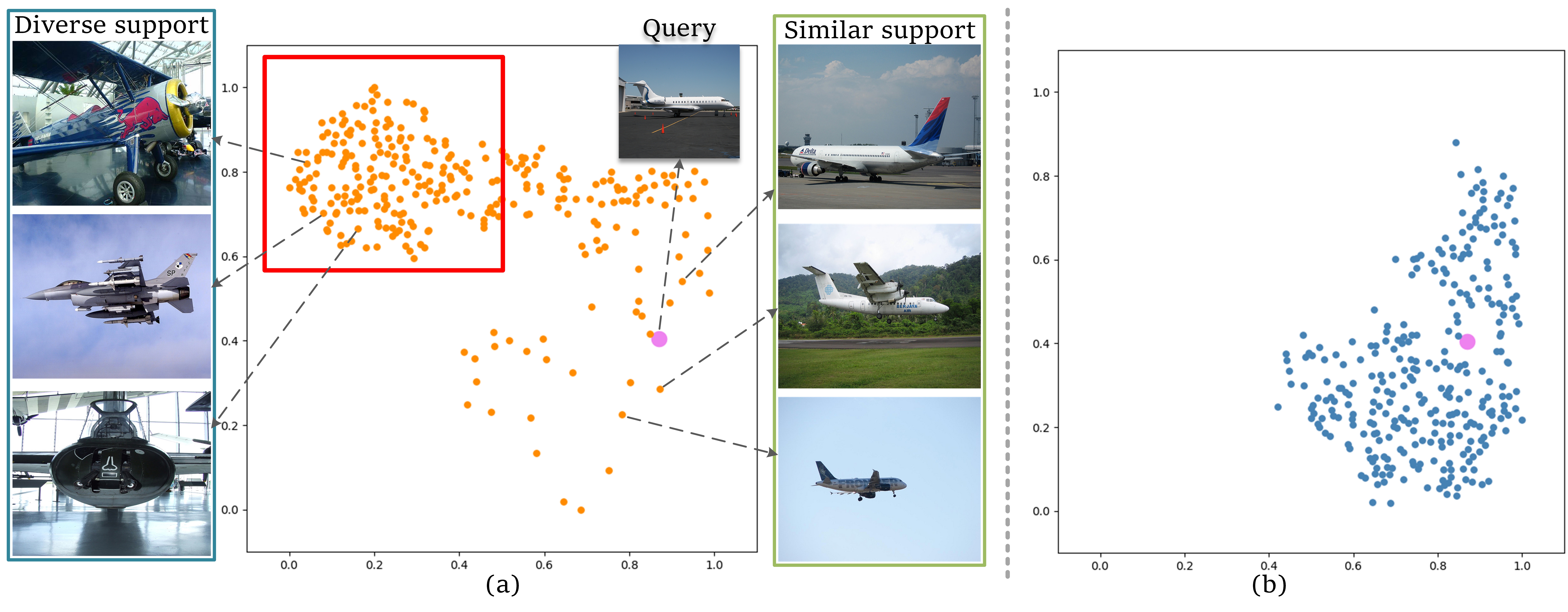}
	\end{center}
	\setlength{\abovecaptionskip}{0.cm}
	\caption{t-SNE visualization of the prototype distribution. (a): The distribution of \textcolor{orange}{support} and \textcolor{magenta}{query} prototypes. (b): The distribution of \textcolor{blue}{intermediate} and \textcolor{magenta}{query} prototypes. Our proposed intermediate prototypes are closer to the query than the support prototypes.}
	\label{fig:mot}
	\vspace{-5mm}
\end{figure*}

However, these methods all rely heavily on the category information extracted from the support set. Although it provides deterministic category information guidance, there may exist inherent intra-class diversity between query and support samples, which is collectively ignore by existing works. 
In Figure~\ref{fig:mot} (a), we show the distribution of some support prototypes (in \textcolor{orange}{orange}) and the prototype generated from a query image (in \textcolor{magenta}{magenta}) of the airplane class. We observe that for the support images that are similar with the query image (marked as ``Similar support'' on the right), their prototypes are close to the query prototype in the feature space, in which case the matching network can work well. However, for the support images that have large diversity in pose and appearance compared with the query (marked as ``Diverse support'' on the left), the distance between the support and query prototypes will be faraway.
% hence the matching network can not handle this gap.
% support images (left) that are far away from query in prototype embedding are named as the hard support, while others (right) that are closer to query are named as the easy support.
% When the objects in the support and query images are similar, the  However, 
% and its extent can be reflected by the distance of feature space . As shown  
In such a case, if we forcibly migrate the category information in the support prototype to the query, a large category information bias is inevitably introduced.

Therefore, our work aims to relieve this problem by introducing an intermediate prototype that could bridge the category information gap between query and support images through our proposed Intermediate Prototype Mining Transformer (IPMT). Each IPMT layer consists of two steps, \ie, Intermediate Prototype Mining (IPM) and Query Activation (QA). In IPM we learn the intermediate prototype via combing both the deterministic category information from the support images and the adaptive category knowledge from the query image. Then, we use the learned prototype to activate the query feature map in QA. Furthermore, our IPMT is used in an iterative way to progressively improve the quality of the learned prototype and the activated query feature.
% fully explore the category information of query to serve the learning of intermediate prototype. 
As such, the intermediate prototype can successfully reduce the category information gap with the query set, which is verified by the distribution in Figure~\ref{fig:mot} (b).
% , and the output of last IPMT layer is the final activated query feature which is used to segment a accurate prediction.

In summary, our main contributions can be concluded as:
1) To the best of our knowledge, this is the first time to focus on the intra-class diversity between support and query in FSS, and we propose the idea of intermediate prototype to relieve the existing category information gap issue.
2) We propose a novel IPMT to explicitly mine the intermediate prototype which contains both the deterministic information from the support set and the adaptive category knowledge from the query.
3) We present an iterative learning scheme to fully explore the intermediate category information hidden in both support and query and update the query feature.
4) Extensive experiments on PASCAL-${5^{i}} $ and COCO-$ 20^{i}$ show that our proposed IPMT brings a significant improvement over state-of-the-art methods.

\vspace{-1.5mm}
\section{Related Work}
\vspace{-1.5mm}
\subsection{Visual Transformer}
\vspace{-1mm}
Recently, transformer is introduced into the field of artificial intelligence and has attracted significant attention from many researchers.  
\cite{vaswani2017attention} first proposes the transformer architecture composed of self-attention and feed-forward layers, and also achieves remarkable performance in natural language processing. Very recently, transformer has been found to be able to obtain excellent results in computer vision. Specifically, 
\cite{dosovitskiy2020image} firstly introduces transformer into vision tasks and proposes vision transformer (ViT) by treating non-overlapped image patches as a series of tokens.
% A class-token is additionally added to interact with the patch tokens for image classification. 
Subsequently, a lot of works are devoted to tailoring the transformer structure to be more suitable for computing vision tasks. \cite{liu2021swin} utilizes shifted window self-attention and patch merging to reduce the computational cost and aggregate features, respectively. \cite{wang2021pyramid} imitates the feature pyramid structure in CNNs and proposes the pyramid vision transformer, which reduces the computational cost of self-attention.
% by pooling keys and values. 

Other transformer-based methods are proposed for various tasks, such as object detection\cite{zhu2020deformable,carion2020end}, semantic segmentation\cite{zheng2021rethinking}, panoptic segmentation\cite{wang2021max}, multiple object tracking\cite{sun2020transtrack} and so on. \cite{carion2020end} proposes an transformer-based end-to-end object detection framework and utilizes object queries to locate the objects. 
% Their method does not rely on many hand-designed components and also outperforms the Faster-RCNN \cite{ren2015faster} baseline.
\cite{guo2021sotr} replaces the multi-head attention with twin attention to interact with the context both on row and column features for instance segmentation. In the above works, learnable Queries\footnote{Here, we use the initially capitalized `Query' and the lowercase `query' to distinguish the context of query in transformer and FSS respectively.} are utilized to aggregate the context-information from feature maps for further regression or segmentation. 
\cite{cheng2021mask2former} introduces the masked attention to extract relevant features instead of global features by limiting the cross-attention regions within predicted masks.
Motivated by these works, we use a learnable prototype as the intermediate prototype to dig out adaptive category information from both query and support images.
\vspace{-1mm}
\subsection{Few-shot Semantic Segmentation}
\vspace{-1mm}
FSS is a natural extension of semantic segmentation in the condition of a few annotated samples. The typical paradigm proposed by SG-one \cite{shaban2017one} is using two-branch networks. A conditioning branch extracts the category context from the support images and another segmentation branch segments the query image under the guidance of the former. Following this paradigm, many approaches \cite{yang2020prototype,li2021adaptive,zhang2021self,liu2021anti,tian2020prior,wu2021learning} are proposed to explore how to fully excavate the category information from the support images. 
% CANet\cite{zhang2019canet} uses a dense comparison module to concatenate the query feature and expanded support feature to activate query features. Then, the activated feature is fed into an iterative optimization module to refine the prediction progressively. 
For mining more abundant category information from the support images, ASGNet\cite{yang2020prototype} and PMM\cite{li2021adaptive} construct multiple prototypes using parameter-free methods, i.e. superpixel-guided clustering and the expectation-maximum algorithm, to cluster the support foreground features into multiple prototypes and then activate different areas in the query image.
SCL\cite{zhang2021self} utilizes the missing parts in the initial segmentation result of the support images to form auxiliary support vectors and then merge them in a cross-guidance module to obtain a better prediction.
In PFENet\cite{tian2020prior}, a prior mask is generated by calculating the cosine-similarity between support and query features in high-level. Then, a feature enrichment module is applied to perform dense comparison on different feature scales obtained by adaptive pooling. 
Furthermore, MMNet\cite{wu2021learning} introduces meta-class memory to store the meta-information during training and applies it into novel classes during the inference stage.
However, all above methods ignore the inherent intra-class differences between query and support images and transfer the support information to the query image forcibly. Our work aims to relieve this problem and propose the intermediate prototype to bridge the category information gap between query and support images.
\vspace{-1mm}
\subsection{Transformer-based Few-shot Semantic Segmentation}
\vspace{-1mm}
 \cite{lu2021simpler} introduces the multi-head attention as an attention module to transfer the classifier weights from support to query. However, it does not take full advantage of the transformer on incorporating long-range dependencies. 
\cite{zhang2021few} proposes a Cycle-Consistent TRansformer (CyCTR) module to 
% aggregate pixel-wise support features into query ones. It 
select relevant pixel-level support features to perform cross-attention with the query feature.
In our work, instead of performing cross-attention between support and query features, we leverage a learnable Query as the intermediate prototype to aggregate the category information from both support and query images and refine the query feature using this prototype.

\vspace{-1.5mm}
\section{Problem Definition}
\vspace{-1.5mm}

In FSS, the whole dataset is divided into two disjoint subsets $ \mathcal D_{train}$  and $ \mathcal D_{test}$ based on the object categories they contain. An FSS model is expected to learn the meta knowledge on $ \mathcal D_{train}$  with sufficient labeled images and generalize to unseen categories on $ \mathcal D_{test}$  with scarce labeled images. Following the previous meta-learning paradigm, we execute the episodic training strategy to train our model. Specifically, these two subsets are both partitioned into numerous episodes, each of which randomly samples $K+1$ image-mask pairs. For one episode, $K$ pairs compose the support set $ \mathcal S=\{(\mathbf{I_i^s},\mathbf{M_i^s})\}_{i=1}^K$ and the rest one pair composes the query set $ \mathcal Q=\{(\mathbf{I^q},\mathbf{M^q})\}$, where $\mathbf{I^*} \in \mathbb{R}$$^{H\times W\times 3} $ and $ \mathbf{M^*} \in \mathbb{R}$$^{H\times W}$ denote the RGB images and their corresponding binary masks, respectively. In each episode sampled from $ \mathcal D_{train}$, the model is trained to predict the query mask supervised by $\mathbf{M^q}$ under the guidance of the support set. Then, the trained model is evaluated on $ \mathcal D_{test}$ to segment unseen categories straightly without any further optimization.

\vspace{-1.5mm}
\section{A Review of Transformer}
\vspace{-1.5mm}
\label{review}
We first review the typical transformer model here. As the form in \cite{vaswani2017attention}, a transformer layer mainly consists of an attention block and a multi-layer perception (MLP) block. The former is used to aggregate global contexts and the latter performs embedding updating.
% Note that layer normalization \cite{ba2016layer} and residual connection are applied for both layers after each of them.

\vspace{-3mm}
\paragraph{Attention Block.}
Given an input token sequence $\mathbf{X} \in \mathbb{R}$$^{L_1\times C} $ and a context token sequence $\mathbf{Y} \in \mathbb{R}$$^{L_2\times C} $, where $L_1,L_2$ are the length of the two sequences and $C$ is the channel dimension of their embeddings, respectively, the attention block first computes the attention weight matrix:
\begin{equation}
    \mathbf{A}(\mathbf{X},\mathbf{Y}) = \frac{\mathbf{X W_q (Y W_k)^{\top}}} {\sqrt{d}}, 
\end{equation}
% where $\mathbf{A} \in \mathbb{R}$$^{HW\times HW}$ and $\sqrt{d}$ is the scale factor related to the channel dimensionality of inputs (\ie, $\mathbf{Q, K}$). $\mathbf{W_q}$ and $\mathbf{W_k}$ are weights from two linear layers to obtain the Query and Key.
% Then, the softmax operation is used to normalize the attention weight $\mathbf{A}$ for further decoupling with input $\mathbf{V}$, which is summarized as:
where $\mathbf{W_q}$ and $\mathbf{W_k}\in \mathbb{R}$$^{C\times d} $ are linear transformation weight matrixes and $\sqrt{d}$ is the scale factor.
Then, $\mathbf{A}(\mathbf{X},\mathbf{Y})$ is normalized and then used to aggregate the global context from $\mathbf{Y}$:
\begin{equation}
\label{eq:att}
    \mathbf{Attn}(\mathbf{X},\mathbf{Y}) = Softmax(\mathbf{A}(\mathbf{X},\mathbf{Y})) \mathbf{Y W_v},
\end{equation}
where $\mathbf{W}_v\in \mathbb{R}$$^{C\times C}$ is another linear transformation weight matrix.
%nian:
% \begin{equation}
%     \label{eq:attention}
%     \mathbf{Attn}(\mathbf{X},\mathbf{Y}) = softmax(\mathbf{W_q X} \mathbf{(W_k Y)}^{\top})\mathbf{W_v Y},
% \end{equation}
% where $\mathbf{W_q}$,$\mathbf{W_k}$, and $\mathbf{W_k}\in \mathbb{R}$$^{C\times C} $ are linear transformation weights.

If $\mathbf{X, Y}$ are the same feature, the attention operation is called self-attention which propagates contexts among different tokens. If they are not the same, it is named cross-attention which conveys relevant information from $\mathbf{Y}$ to $\mathbf{X}$. Usually multi-head attention \cite{vaswani2017attention} is used to boost the model performance.

\vspace{-3mm}
\paragraph{Multi-layer Perception Block.}
After the attention block, a MLP block is applied to each token separately and identically to further transform the token embeddings. Specifically, MLP is implemented using two linear projection layers with a ReLU activation in between. Given a token sequence $\mathbf{X} \in \mathbb{R}$$^{L_1\times C} $ as the input, it is formulated as:
\begin{equation}
        \mathbf{MLP}(\mathbf{X}) = ReLU(\mathbf{X W_1 + b_1})\mathbf{W_2 + b_2},
\end{equation}
where $\mathbf{W_*}$ and $\mathbf{b_*}$ denote the linear transformation weight matrixes and biases, respectively. We follow \cite{vaswani2017attention} and set the channel dimensions of the first and the second layer to be $4C$ and $C$, respectively. Note that layer normalization \cite{ba2016layer} and residual connections are 
% also applied for both attention and MLP blocks after each of them but are 
omitted here for simplicity.

% \begin{figure*}[thbp]
% 	\begin{center}
% 		\includegraphics[scale = 0.8]{figure/ipmt.jpg}
% 	\end{center}
% 	\setlength{\abovecaptionskip}{0.cm}
% 	\caption{Overall architecture of the proposed method for few-shot semantic segmentation. Our network is composed of four parts. After extracting features from both the support and query images via a pre-trained backbone, our Background Mining Modul (BGMM) is performed to obtain a BG prototype and segment the BG regions. Meanwhile, Background Eliminating Module (BGEM) is performed to eliminate the BG regions. The third part is to obtain the activated query feature and further an initial target prediction via Feature Matching (FM). The last part is to eliminate the distracting objects by our proposed Distracting Objects Eliminating Module (DOEM).}
% 	\label{fig:ipmt}
% 	\vspace{-3mm}
% \end{figure*}

\vspace{-1.5mm}
\section{Intermediate Prototype Mining Transformer}
\begin{figure*}[htbp]
	\begin{center}
		\includegraphics[width = 0.95\linewidth]{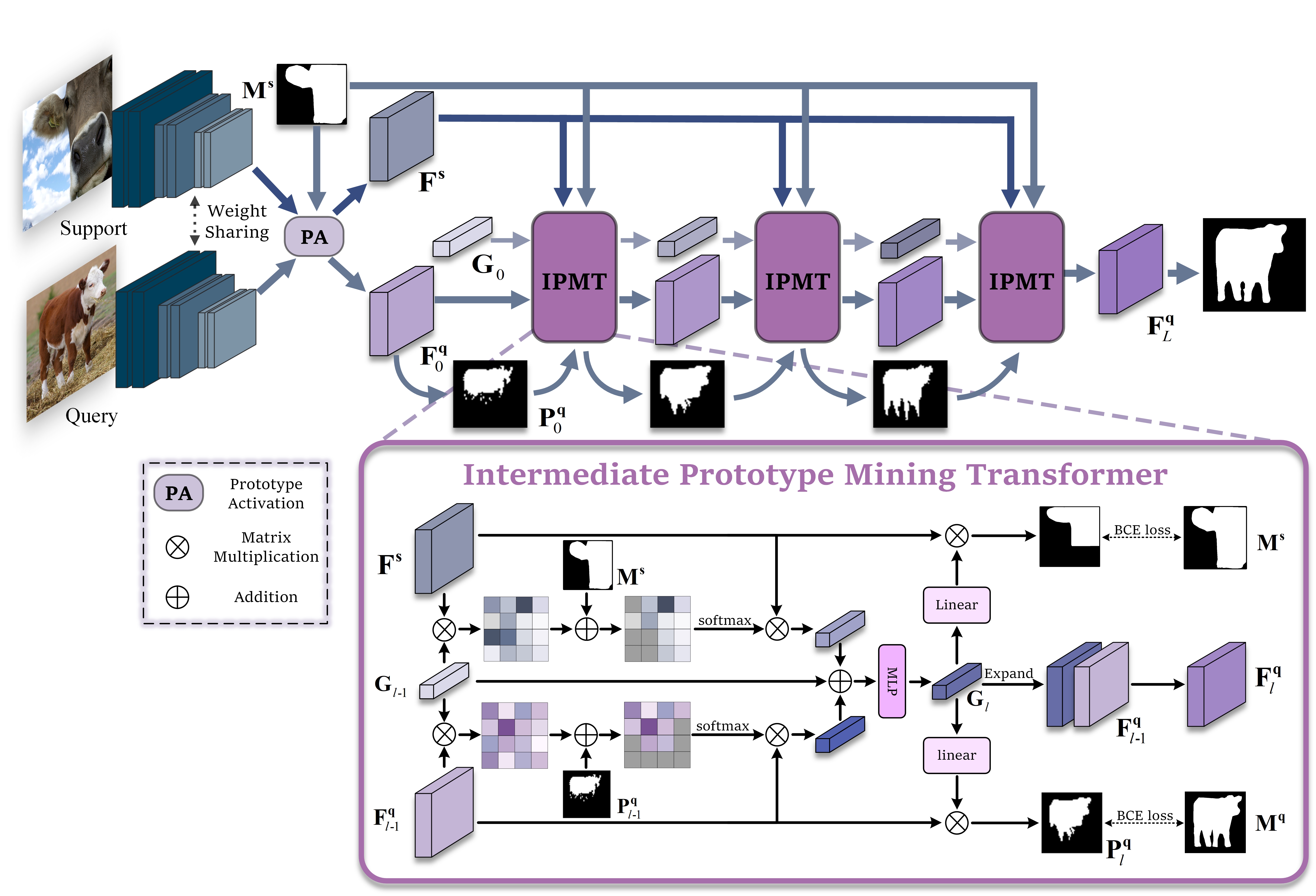}
	\end{center}
	\setlength{\abovecaptionskip}{0.cm}
	\caption{Overall architecture of our proposed IPMT. Support and query images are first fed into two pre-trained backbone encoders to extract features. Then, we follow previous works \cite{tian2020prior,zhang2021few} and conduct prototype activation (PA) to obtain the support and initial query features $\mathbf{F^s}$ and $\mathbf{F}_{0}^{\mathbf{q}}$, respectively. Meanwhile, the initial query segmentation mask $\mathbf{P}_{0}^{\mathbf{q}}$ is obtained from $\mathbf{F}_{0}^{\mathbf{q}}$. Next, we feed $\mathbf{F^s}$, $\mathbf{F}_{0}^{\mathbf{q}}$, $\mathbf{P}_{0}^{\mathbf{q}}$, and the support mask $\mathbf{M}^{s}$, the initial intermediate prototype $\mathbf{G}_0$ into our IPMT layers to iteratively update the prototype, the query mask, and the query feature. After $L$ iterations, the final query feature $\mathbf{F}_{L}^{\mathbf{q}}$ is used to obtain the final query segmentation result. }
	\label{fig:overall}
	\vspace{-3mm}
\end{figure*}

\vspace{-1.5mm}
We now present our proposed Intermediate Prototype Mining Transformer (IPMT), as Figure~\ref{fig:overall} shown, for few-shot semantic segmentation. Each IPMT layer consists of two steps, \ie, Intermediate Prototype Mining (IPM) and Query Activation (QA). IPM is used to mine the intermediate prototype from both support and query features while QA is designed to activate the query feature map using the learned prototype. We adopt a duplex segmentation loss (DSL) to supervise the learning of the intermediate prototype in each IPMT layer. Furthermore, we propose to perform the intermediate prototype mining in an iterative way, thus boosting the quality of the learned prototype and the segmentation results progressively. Next, we will describe them in details.

\vspace{-1mm}
\subsection{Intermediate Prototype Mining}
\vspace{-1mm}
\label{decoder}
Our IPM has a learnable prototype
%Because the MLP is detailed in Section \ref{review}, in this section, we do not reintroduce it and pay more attention to the key component, \ie, MMA.
%
%\paragraph{Learnable Prototype.}
%Because the inherent intra-class differences between query and support feature map, it is unreliable to obtain the category information from support alone.
$\mathbf{G}$ to extract adaptive category information from both query and support images using Masked Attention (MA). 
%explicitly, instead of explicitly selecting features from support.
%Specifically, following the content feature in DETR-style module \cite{zhu2020deformable,carion2020end,dai2021up,meng2021conditional}, we introduce a learnable prototype $\mathbf{G}$ to aggregate the category information after iterations of Transformer, 
The prototype $\mathbf{G} \in \mathbb{R}$$^{1\times C}$ is initially a category- and image-agnostic vector that encodes general segmentation prior and will be updated by MA in each episode, encoding adaptive category information for the target category in that episode.

\paragraph{Masked Attention.} We leverage cross-attention to update $\mathbf{G}$ by using both support and query features as the context. Furthermore, to make $\mathbf{G}$ only focus on the target regions and extract the category information without noises, we follow \cite{cheng2021mask2former} and use support and query masks to limit the attended region in the attention matrix. Specifically, given a flattened support or query feature $\mathbf{F} \in \mathbb{R}$$^{hw\times C}$ and a corresponding binary segmentation mask $\mathbf{P} \in \mathbb{R}$$^{h\times w}$, we first compute the attention weight matrix $\mathbf{A}(\mathbf{G},\mathbf{F}) \in \mathbb{R}$$^{1\times hw}$. 
% \begin{equation}
%     \mathbf{\hat{A}} = \frac{\mathbf{W_G G (W_k x)^T}} {\sqrt{d}} 
% \end{equation}
% where, $\mathbf{\hat{A}} \in \mathbb{R}$$^{1\times HW}$ and $\sqrt{d}$ is the scale factor related to the channel dimensionality of $\mathbf{x}$. $\mathbf{W_G}$ and $\mathbf{W_k}$ are weights from two linear layers to obtain the Query and Key. 
Then, an attention mask is computed by following \cite{cheng2021mask2former}: 
\begin{equation}
\mathbf{\hat{P}}(i) =\left\{
\begin{array}{rcl}
0 & & \text{if} \ \mathbf{P}\text{(i)=1}\\
{- \infty} & & \text{otherwise}
\end{array} \right.,
\end{equation}
where $i$ denotes the location index. Next, we use $\mathbf{\hat{P}}$ to modulate the attention weights, leading to the masked attention:
\begin{equation}
    \mathbf{MaskAttn}(\mathbf{G},\mathbf{F},\mathbf{P}) = Softmax(\mathbf{A}(\mathbf{G},\mathbf{F})+\mathbf{Vec(\hat{P})}) \mathbf{F W_v},
\end{equation}
where $\mathbf{Vec(\cdot)}$ is the vectorization operation. As such, the normalized attention weights on the background regions are close to zero, making sure the prototype only be updated by relevant contexts of the desired category.

In our IPM, on one hand, we use the support feature $\mathbf{F^s}$ and the ground-truth support mask $\mathbf{M^s}$ to update $\mathbf{G}$, providing deterministic category information since $\mathbf{M^s}$ is definitely accurate. On the other hand, we also leverage the query feature $\mathbf{F^q}$ and a query prediction mask $\mathbf{P^q}$ to provide query-adaptive category knowledge for $\mathbf{G}$, thus reducing the category information gap between support and query images. After that, an MLP block is further used on the learned prototype. The whole process can be formulated as:
% \begin{footnotesize}
\begin{equation}
\small{
\label{IPM}
    \mathbf{IPM}(\mathbf{G},\mathbf{F^s},\mathbf{F^q},\mathbf{M^s},\mathbf{P^q}) =  \mathbf{MLP}(\mathbf{MaskAttn}(\mathbf{G},\mathbf{F^s},\mathbf{M^s}) +\mathbf{MaskAttn}(\mathbf{G},\mathbf{F^q},\mathbf{P^q})+\mathbf{G}).
}
\end{equation}
% \end{footnotesize}
% Note that $\mathbf{M^s}$ is the ground-truth of support, and $\mathbf{P^q}$ is the initial prediction mask of query.
% binarized query prediction of the previous $(l-1)$ IPTM layer via setting the thresholded at 0.5 after sigmoid operation. 
Please note that the two masked attention operations do \emph{not} share weights since the two segmentation masks have different uncertainty. For the $K$-shot setting, we simply average the outputs of $\mathbf{MaskAttn}$ on support.

\subsection{Query Activation}
In this step, QA is used to activate the target regions in the query feature map $\mathbf{F^q}$ under the guidance of the learned prototype $\mathbf{G}$. Previous works \cite{tian2020prior,zhang2019canet} have demonstrated that it is an essential operation for FSS to pass the category information to the query feature map and provide specific segmentation cues. Specifically, $\mathbf{G}$ is expanded and concatenated with $\mathbf{F^q} \in \mathbb{R}$$^{h\times w\times C}$ to activate the target regions:
\begin{equation}
    \mathbf{QA(G,F^q)} = \mathcal{F}_{actv} (\mathbf{G \circledcirc F^q }),
\end{equation}
where $\circledcirc $ represents the concatenation operation, and $\mathcal{F}_{actv}$ is a simple activation network which consists of a $1\times1$ convolutional layer, a ReLU layer, and a $3\times3$ convolutional layer. Additionally, we also follow \cite{zhang2021few} to use a multi-head deformable self-attention layer \cite{zhu2020deformable} for further aggregating context information in the query feature.

% \vspace{-1mm}
\subsection{Duplex Segmentation Loss}
% \vspace{-1mm}
\label{mgm}
To facilitate the learning of the adaptive category information in $\mathbf{G}$, we use it to generate two segmentation masks on both support and query images and calculate two segmentation losses. Specifically, motivated by \cite{cheng2021mask2former}, we use $\mathbf{G}$ to generate a mask embedding and then conduct multiplication with the image feature maps for obtaining segmentation masks. The mask generation (MG) process is formulated as:
\begin{eqnarray}
        & \mathbf{MG(G,F^q)} =   Sigmoid(\mathbf{G W_m} \mathbf{(F^q)^{\top}})  ,\\
        & \mathbf{MG(G,F^s)} =   Sigmoid(\mathbf{G W_m} \mathbf{(F^s)^{\top}}) ,
\end{eqnarray}
where $\mathbf{W_m} \in \mathbb{R}$$^{C\times C}$ is a linear projection weight matrix for generating the mask embedding.

Next, the standard binary cross-entropy (BCE) loss is calculated between the generated masks and the ground truth, \ie, $\mathbf{M^q}$ and $\mathbf{M^s}$, as our duplex segmentation loss to optimize the prototype learning process:
\begin{equation}
    \mathcal{L}^{dsl} = \alpha BCE\mathbf{(\mathbf{MG(G,F^q)}, M^q)} + (1-\alpha) BCE\mathbf{(\mathbf{MG(G,F^s)}, M^s)}.
\end{equation}
Here, $\alpha$ is a hyperparameter to balance the losses between query and support predictions.
% \vspace{-1mm}
\subsection{Iterative Prototype Mining}
% \vspace{-1mm}
Since one IPMT layer can update the intermediate prototype $\mathbf{G}$, the query feature map $\mathbf{F^q}$, and the query segmentation mask $\mathbf{P^{q}}$,
% given the input of the initial prototype $\mathbf{G}$ and the two feature maps and two segmentation masks, . Hence, 
we can iteratively perform this process and obtain better and better $\mathbf{G}$ and $\mathbf{F^q}$, finally making the segmentation results effectively boosted. Suppose we have $L$ iterative IPMT layers, then for each layer $l$ we have:
%Next, the randomly initialized intermediate prototype, the query feature map, the initial prediction, the support feature, and the support ground-truth mask are fed into our model (\ie, IPMT) to update the query feature and predict masks iteratively, which is summarized as:
\begin{equation}
        \mathbf{G}_l,\mathbf{F}_{l}^{\mathbf{q}},\mathbf{P}_{l}^{\mathbf{q}} = \mathbf{IPMT}(\mathbf{G}_{l-1},\mathbf{F^s},\mathbf{F}_{l-1}^{\mathbf{q}},\mathbf{M^s},\mathbf{P}_{l-1}^{\mathbf{q}}), 
\end{equation}
which can be broken down into the following steps:
\begin{gather}
    \mathbf{G}_l  =  \mathbf{IPM}(\mathbf{G}_{l-1},\mathbf{F^s},\mathbf{F}_{l-1}^{\mathbf{q}},\mathbf{M^s},\mathbf{P}_{l-1}^{\mathbf{q}}),\\  
    \mathbf{F}_{l}^{\mathbf{q}}= \mathbf{QA}(\mathbf{G}_l,\mathbf{F}_{l-1}^{\mathbf{q}}),\\
    \mathbf{P}_{l}^{\mathbf{q}} = \mathbf{MG}(\mathbf{G}_l,\mathbf{F}_{l-1}^{\mathbf{q}})\geq 0.5.
\end{gather}
Here, since the masked attention requires a binary mask as the input, we use 0.5 as the threshold to generate $\mathbf{P}_{l}^{\mathbf{q}}$.

% Additionally, to ensure the intermediate prototype aggregate the category information from both query and support,  the BCE Loss is calculated between $\mathbf{p^q_l}$ with $\mathbf{M^q}$ and $\mathbf{p^s_l}$ with $\mathbf{M^s}$ respectively to optimize predictions:
% \begin{equation}
%     \mathcal{L}^{de}_l = \alpha BCE\mathbf{(p^q_l, M^q)} + (1-\alpha) BCE\mathbf{(p^s_l, M^s)}
% \end{equation}
% where, $\alpha$ is a hyperparameter to balance the loss between query and support predictions.
%After iterations, the final activated query feature $\mathbf{F^q_{L}}$ is used to predict the final result, where $L$ is the number of IPMT layers. Dice loss is used as the primary loss to optimize the final prediction.
% \vspace{-1.5mm}
\section{Few-shot Semantic Segmentation Model}
% \vspace{-1.5mm}

Following previous works \cite{tian2020prior,zhang2021few}, we input query and support images into a fixed and shared encoder backbone such as the ResNet family \cite{he2016deep} to obtain multi-level features. Then, we concatenate the outputs of the third and fourth encoder blocks together and then adopt a $1\times1$ convolutional layer to generate middle-level query and support features, respectively. We also calculate the similarity between the high-level query and support features at the fifth encoder block to produce a prior mask and use masked average pooling on the support feature map to obtain a support prototype. Next, the query feature map, prior mask, and the expanded prototype are concatenated and transformed using a $1\times1$ convolutional layer to obtain the initial query feature $\mathbf{F}_{0}^{\mathbf{q}}$. We also concatenate the middle-level support feature with the expanded prototype to generate the support feature $\mathbf{F^s}$. All the above processes are common methods in FSS and termed prototype activation (PA) in Figure~\ref{fig:overall}. For more details please refer to \cite{tian2020prior}.

As for the iterative learning, we feed $\mathbf{F}_{0}^{\mathbf{q}}$ into two convolutional layers to obtain the initial query segmentation prediction $\mathbf{P}_{0}^{\mathbf{q}}$. 
% The simple segmentation network consists of a 1 × 1 convolutional layer, a ReLU layer, and a 3 × 3 convolutional layer. 
The initial intermediate prototype $\mathbf{G}_0$ is randomly initialized at the beginning of the training and then optimized on the training set. Next, we feed $\mathbf{G}_{0},\mathbf{F^s},\mathbf{F}_{0}^{\mathbf{q}},\mathbf{M^s}$, and $\mathbf{P}_{0}^{\mathbf{q}}$ into our iterative IPMT layers to perform intermediate prototype mining. After $L$ iterations, the final activated query feature $\mathbf{F}_{L}^{\mathbf{q}}$ is used to predict the final segmentation result via two convolutional layers.
% which consists of a 1 × 1 convolutional layer, a ReLU layer, and a 3 × 3 convolutional layer. 
The dice loss \cite{milletari2016v} is used here to optimize the training of the whole model.

% \vspace{-1.5mm}
\section{Experiments}
% \vspace{-1.5mm}

\begin{table*}[]
	\centering
	\caption{Class mIoU results of four folds on PASCAL-$5^{i}$. The results of ‘Mean’ are the averaged class mIoU scores of all four folds. \textbf{\textcolor{red}{Red}}/\textbf{\textcolor{blue}{Blue}} indicates the best/$2^{nd}$ results.}
	\label{Voccompar}
	\resizebox{\linewidth}{!}{
\begin{tabular}{@{}c|ccccccccccc@{}}
\toprule
                             & \multicolumn{1}{c|}{}                                   & \multicolumn{5}{c|}{1-shot}                                                                                                                                                    & \multicolumn{5}{c}{5-shot}                                                                                                                 \\ %\cmidrule(l){3-12} 
\multirow{-2}{*}{Backbone}   & \multicolumn{1}{c|}{\multirow{-2}{*}{Methods}}          & Fold-0                       & Fold-1                       & Fold-2                       & Fold-3                       & \multicolumn{1}{c|}{ \textbf{Mean}}                         & Fold-0                    & Fold-1                    & Fold-2                    & Fold-3                    &  \textbf{Mean}                       \\ \midrule
                             & \multicolumn{1}{c|}{RPMMs(ECCV'20)\cite{yang2020prototype}     }                     & 55.2                         & 66.9                         & 52.6                         & 50.7                         & \multicolumn{1}{c|}{56.3}                         & 56.3                      & 67.3                      & 54.5                      & 51.0                      & 57.3                       \\
                             & \multicolumn{1}{c|}{PFENet(TPAMI'20)}                   & 61.7                         & 69.5                         & 55.4                         & 56.3                         & \multicolumn{1}{c|}{60.8}                         & 63.1                      & 70.7                      & 55.8                      & 57.9                      & 61.9                       \\
                             & \multicolumn{1}{c|}{RePRI(CVPR'21)\cite{boudiaf2021few}}                     & 59.8                         & 68.3                         & \textbf{\textcolor{red}{62.1}}                         & 48.5                         & \multicolumn{1}{c|}{59.7}                         & 64.6                      & 71.4                      & \textbf{\textcolor{red}{71.1}}                      & 59.3                      & 66.6                       \\
                             & \multicolumn{1}{c|}{HSNet(ICCV'21)\cite{min2021hypercorrelation}}                     & 64.3                         & 70.7                         & 60.3                         & \textbf{\textcolor{blue}{60.5}}                         & \multicolumn{1}{c|}{64.0}                         & \textbf{\textcolor{blue}{70.3}}                      & 73.2                      & 67.4                      & \textbf{\textcolor{red}{67.1}}                      & \textbf{\textcolor{red}{69.5}}                       \\
                             & \multicolumn{1}{c|}{CWT(ICCV'21)\cite{lu2021simpler}}                       & 56.3                         & 62.0                         & 59.9                         & 47.2                         & \multicolumn{1}{c|}{56.4}                         & 61.3                      & 68.5                      & \textbf{\textcolor{blue}{68.5}}                      & 56.6                      & 63.7                       \\
                             & \multicolumn{1}{c|}{CyCTR(NeurIPS'21)\cite{zhang2021few}}                  & \textbf{\textcolor{blue}{65.7}}                         & 71.0                         & 59.5                         & 59.7                         & \multicolumn{1}{c|}{64.0}                         & 69.3                      & \textbf{\textcolor{blue}{73.5}}                     & 63.8                      & 63.5                      & 67.5                       \\
                             & \multicolumn{1}{c|}{NERTNet(CVPR'22)\cite{liu2022learning}}                  & 65.4                     & \textbf{\textcolor{blue}{72.3}}                        & 59.4                         & 59.8                        & \multicolumn{1}{c|}{\textbf{\textcolor{blue}{64.2}}}                         & 66.2                      & 72.8                      & 61.7                      & 62.2                      & 65.7                       \\
                             & \multicolumn{1}{c|}{DCP(IJCAI'22)\cite{lang2022beyond}}                      & 63.8                         & 70.5                         & \textbf{\textcolor{blue}{61.2}}                         & 55.7                         & \multicolumn{1}{c|}{62.8}                         & 67.2                      & 73.1                      & 66.4                      & \textbf{\textcolor{blue}{64.5}}                      & 67.8                       \\

\multirow{-9}{*}{ResNet-50}  & \multicolumn{1}{c|}{\cellcolor[HTML]{C0C0C0}IPMT(ours)} & \cellcolor[HTML]{C0C0C0}\textbf{\textcolor{red}{72.8}} & \cellcolor[HTML]{C0C0C0}\textbf{\textcolor{red}{73.7}} & \cellcolor[HTML]{C0C0C0}59.2 & \cellcolor[HTML]{C0C0C0}\textbf{\textcolor{red}{61.6}} & \multicolumn{1}{c|}{\cellcolor[HTML]{C0C0C0}\textbf{\textcolor{red}{66.8}}} & \cellcolor[HTML]{C0C0C0}\textbf{\textcolor{red}{73.1}} & \cellcolor[HTML]{C0C0C0}\textbf{\textcolor{red}{74.7}} & \cellcolor[HTML]{C0C0C0}61.6 & \cellcolor[HTML]{C0C0C0}63.4 & \cellcolor[HTML]{C0C0C0}\textbf{\textcolor{blue}{68.2}}  \\ \midrule
                             & \multicolumn{1}{c|}{DAN(ECCV'20)\cite{wang2020few}}                       & 54.7                         & 68.6                         & 57.8                         & 51.6                         & \multicolumn{1}{c|}{58.2}                         & 57.9                      & 69.0                      & 60.1                      & 54.9                      & 60.5                       \\
                             & \multicolumn{1}{c|}{PFENet(TPAMI'20)\cite{tian2020prior}}                   & 60.5                         & 69.4                         & 54.4                         & 55.9                         & \multicolumn{1}{c|}{60.1}                         & 62.8                      & 70.4                      & 54.9                      & 57.6                      & 61.4                       \\
                             & \multicolumn{1}{c|}{CWT(ICCV'21)\cite{lu2021simpler}}                       & 56.9                         & 65.2                         & \textbf{\textcolor{red}{61.2}}                         & 48.8                         & \multicolumn{1}{c|}{58.0}                         & 62.6                      & 70.2                      &\textbf{\textcolor{red}{ 68.8}}                      & 57.2                      & 64.7                       
                             \\
                             & \multicolumn{1}{c|}{NERTNet(CVPR'22)\cite{liu2022learning}}                  & 65.5                     & 71.8                         & \textbf{\textcolor{blue}{59.1}}                         & 58.3                       & \multicolumn{1}{c|}{63.7}                       & 67.9                      & 73.2                      & \textbf{\textcolor{blue}{60.1}}                      & \textbf{\textcolor{red}{66.8}}                      & \textbf{\textcolor{blue}{67.0}}                       \\
                             & \multicolumn{1}{c|}{CyCTR(NeurIPS'21)\cite{zhang2021few}}                  & \textbf{\textcolor{blue}{69.3}}                         & \textbf{\textcolor{blue}{72.7}}                         & 56.5                         & \textbf{\textcolor{blue}{58.6}}                         & \multicolumn{1}{c|}{\textbf{\textcolor{blue}{64.3}}}                         & \textbf{\textcolor{blue}{73.5}}                      & \textbf{\textcolor{blue}{74.0}}                      & 58.6                      & 60.2                      & 66.6                       \\
 
\multirow{-6}{*}{ResNet-101} & \cellcolor[HTML]{C0C0C0}IPMT(ours)                      & \cellcolor[HTML]{C0C0C0}\textbf{\textcolor{red}{71.6}} & \cellcolor[HTML]{C0C0C0}\textbf{\textcolor{red}{73.5}} & \cellcolor[HTML]{C0C0C0}58.0 & \cellcolor[HTML]{C0C0C0}\textbf{\textcolor{red}{61.2}} & \cellcolor[HTML]{C0C0C0}\textbf{\textcolor{red}{66.1}}                      & \cellcolor[HTML]{C0C0C0}\textbf{\textcolor{red}{75.3}} & \cellcolor[HTML]{C0C0C0}\textbf{\textcolor{red}{76.9}} & \cellcolor[HTML]{C0C0C0}59.6 & \cellcolor[HTML]{C0C0C0}\textbf{\textcolor{blue}{65.1}} & \cellcolor[HTML]{C0C0C0}\textbf{\textcolor{red}{69.2}} \\ \bottomrule
\end{tabular}}
\vspace{-5mm}
\end{table*}

\subsection{Datasets and Evaluation Metrics}
% \vspace{-1mm}
\label{data}
\paragraph{Datasets.}
To make a fair comparison with previous works, our model is evaluated on two few-shot semantic segmentation benchmark datasets, \ie, the PASCAL-$5^{i}$ dataset\cite{shaban2017one} and the COCO-$20^{i}$ dataset \cite{nguyen2019feature}. PASCAL-$5^{i}$ is constructed based on the PASCAL VOC 2012 dataset \cite{everingham2010pascal} and additional annotations from SDS \cite{hariharan2011semantic}. It contains 20 categories in total and these categories are partitioned into four folds as in \cite{wang2019panet} for cross validation, where each fold contains five categories. COCO-$20^{i}$ is a larger datasets based on the MSCOCO \cite{lin2014microsoft} dataset. Similar to the division in PASCAL-$5^{i}$, the 80 categories in MSCOCO are also partitioned into four folds for cross validation, where each fold includes 20 categories. For both datasets, we train our model on three folds and evaluate it on the remaining one fold and perform cross validation.

% \vspace{-3mm}
\paragraph{Evaluation Metrics.}
Following previous methods \cite{shaban2017one,siam2019amp,liu2020crnet,liu2020part}, we adopt the class mean intersection over union (mIoU) as a primary evaluation metric. In addition, we also report the results of foreground-background IoU (FB-IoU) for comparison.
% \vspace{-1mm}
\subsection{Implementation Details}
% \vspace{-1mm}
\label{Implementation}
Following previous works, we adopt ResNet-50 and ResNet-101 \cite{he2016deep} as our encoder backbone. Note that they are initialized by the weights pre-trained on ImageNet \cite{russakovsky2015imagenet} and froze during training.

Our proposed IPMT is implemented using PyTorch \cite{paszke2019pytorch} and all the experiments are conducted on one NVIDIA RTX 3090 GPU. We use the same data augmentation setting as \cite{tian2020prior} for fair comparisons. Our model is trained for 200 epochs on PASCAL-$5^{i}$ while 50 epochs on COCO-$20^{i}$, respectively, with the batchsize set to 4. Two optimizers (\ie, SGD and AdamW) are used to train our model. The former is used to optimize the convolutional layers with the initial learning rate, weight decay, and momentum set to $2.5\times10^{-3}$, 0.0001, and 0.9, respectively. The latter is used to optimize the transformer layers by setting the learning rate to $1\times10^{-4}$ and the weight decay to $1\times10^{-2}$. We also use the polynomial annealing policy with the power set to 0.9 to decay the learning rate. For hyper-parameters in our IPMT, the number of 
% points of deformable attention, 
multi-heads and the channel dimension of the image features are set to 8 and 256, respectively. The weight $\alpha$ in DSL is set to 0.3 since a larger weight should be given for the more reliable category knowledge from the support images. During the evaluation, we follow \cite{tian2020prior} to randomly sample 1000 support-query pairs on PASCAL-$5^{i}$ and 4000 pairs on COCO-$20^{i}$, respectively.
\subsection{Comparison with State-of-the-art Methods}
% \vspace{-1mm}
\begin{figure*}[thbp]
	\begin{center}
		\includegraphics[scale = 0.03]{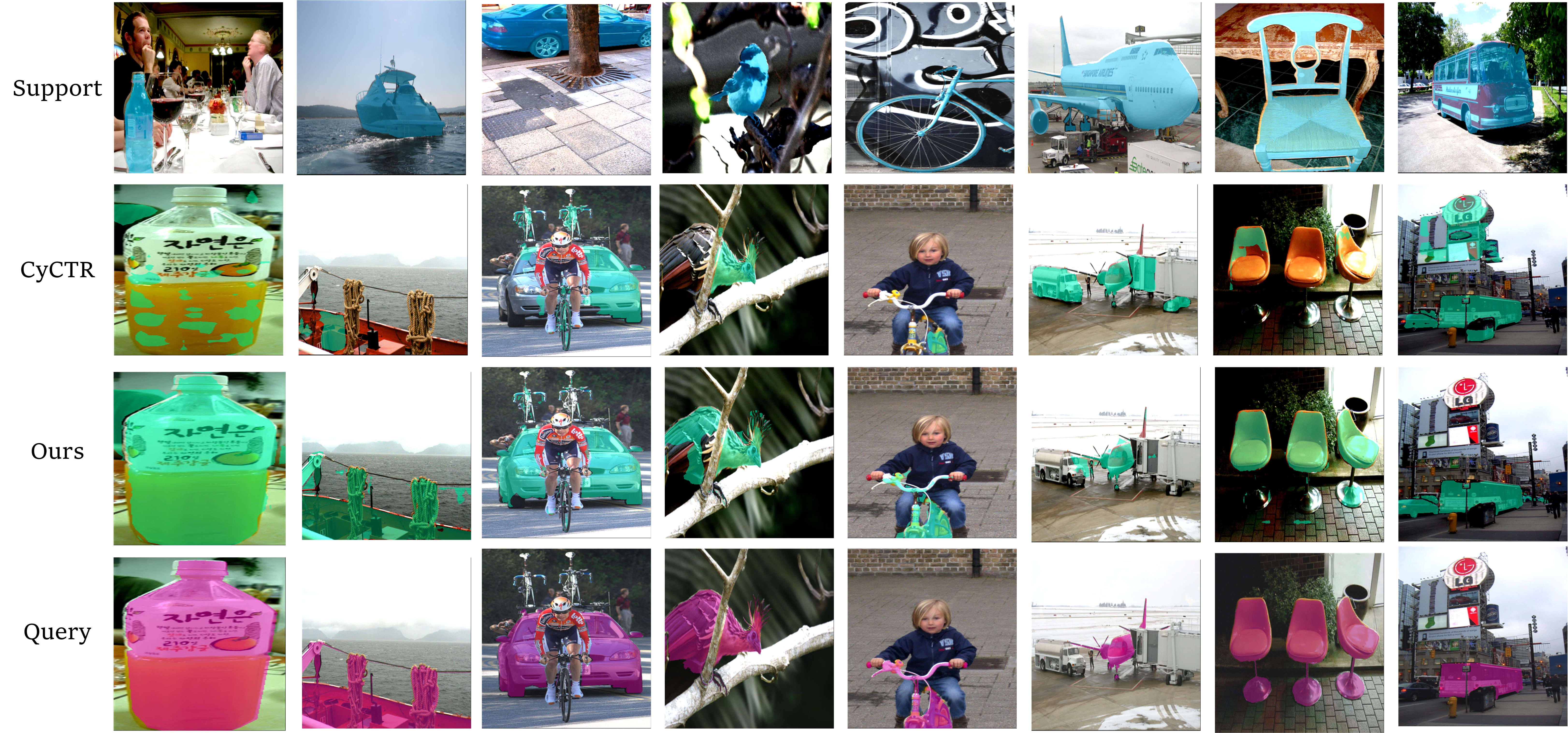}
	\end{center}
	\setlength{\abovecaptionskip}{0.cm}
	\caption{Qualitative comparison between our proposed IPMT and CyCTR \cite{zhang2021few}. From top to bottom: support images, prediction of CyCTR, prediction of IPMT, query images.}
	\label{fig:compar}
	\vspace{-3mm}
\end{figure*}

\begin{table*}[]
	\centering
	\caption{Class mIoU results of four folds on COCO-$20^{i}$. The results of ‘Mean’ are the averaged class mIoU scores of all the four folds. \textbf{\textcolor{red}{Red}}/\textbf{\textcolor{blue}{Blue}} indicates the best/$2^{nd}$ results.
	}
	\label{Cococompar}
	\resizebox{\linewidth}{!}{
\begin{tabular}{@{}c|c|ccccc|ccccc@{}}
\toprule
                             &                                    & \multicolumn{5}{c|}{1-shot}                                                                                                               & \multicolumn{5}{c}{5-shot}                                                                                                                 \\
\multirow{-2}{*}{Backbone}   & \multirow{-2}{*}{Methods}          & Fold-0                    & Fold-1                    & Fold-2                    & Fold-3                    & \textbf{Mean}                      & Fold-0                    & Fold-1                    & Fold-2                    & Fold-3                    & \textbf{Mean}                       \\ \midrule
                             & RPMMs(ECCV'20)\cite{yang2020prototype}                     & 29.5                      & 36.8                      & 29.0                      & 27.0                      & 30.6                      & 33.8                      & 42.0                      & 33.0                      & 33.3                      & 35.5                       \\
                             & RePRI(CVPR'21)\cite{boudiaf2021few}                     & 31.2                      & 38.1                      & 33.3                      & 33.0                      & 34.0                      & 38.5                      & 46.2                      & 40.0                      & 43.6                      & 42.1                       \\
                             & HSNet(ICCV'21)\cite{min2021hypercorrelation}                     & 36.3                      & 43.1                      & 38.7                      & 38.7                      & 39.2                      & 43.3                      & \textbf{\textcolor{red}{51.3}}                      & \textbf{\textcolor{blue}{48.2}}                      & 45.0                      & \textbf{\textcolor{blue}{46.9 }}                      \\
                             & CWT(ICCV'21)\cite{lu2021simpler}                       & 32.2                      & 36.0                      & 31.6                      & 31.6                      & 32.9                      & 40.1                      & 43.8                      & 39.0                      & 42.4                      & 41.3                       \\
                             & CyCTR(NeurIPS'21)\cite{zhang2021few}                  & 38.9                      & 43.0                      & 39.6                      & \textbf{\textcolor{blue}{39.8}}                      & 40.3                      & 41.1                      & 48.9                      & 45.2                      & \textbf{\textcolor{blue}{47.0}}                      & 45.6                       \\
                             & DCP(IJCAI'22)\cite{lang2022beyond}                      & \textbf{\textcolor{blue}{40.9}}                      & \textbf{\textcolor{blue}{43.8}}                      & \textbf{\textcolor{blue}{42.6}}                      & 38.3                      & \textbf{\textcolor{blue}{41.4}}                      & \textbf{\textcolor{red}{45.8}}                      & 49.6                      & 43.7                      & 46.6                      & 46.5                       \\
                             & NERTNet(CVPR'22) \cite{liu2022learning}                 & 36.8                      & 42.6                      & 39.9                      & 37.9                      & 39.3                      & 38.2                      & 44.1                      & 40.4                      & 38.4                      & 40.3                       \\ 
\multirow{-8}{*}{ResNet-50}  & \cellcolor[HTML]{C0C0C0}IPMT(ours) & \cellcolor[HTML]{C0C0C0}\textbf{\textcolor{red}{41.4}} & \cellcolor[HTML]{C0C0C0}\textbf{\textcolor{red}{45.1}} & \cellcolor[HTML]{C0C0C0}\textbf{\textcolor{red}{45.6}} & \cellcolor[HTML]{C0C0C0}\textbf{\textcolor{red}{40.0}} & \cellcolor[HTML]{C0C0C0}\textbf{\textcolor{red}{43.0}} & \cellcolor[HTML]{C0C0C0}\textbf{\textcolor{blue}{43.5}} & \cellcolor[HTML]{C0C0C0}\textbf{\textcolor{blue}{49.7}} & \cellcolor[HTML]{C0C0C0}\textbf{\textcolor{red}{48.7}} & \cellcolor[HTML]{C0C0C0}\textbf{\textcolor{red}{47.9}} & \cellcolor[HTML]{C0C0C0}\textbf{\textcolor{red}{47.5}}  \\ \midrule
                             & PFENet(TPAMI'20)\cite{tian2020prior}                   & 34.3                      & 33.0                      & 32.3                      & 30.1                      & 32.4                      & 38.5                      & 38.6                      & 38.2                      & 34.3                      & 37.4                       \\
                             & CWT(ICCV'21)\cite{lu2021simpler}                       & 30.3                      & 36.6                      & 30.5                      & 32.2                      & 32.4                      & 38.5                      & 46.7                      & 39.4                      & \textbf{\textcolor{blue}{43.2}}                      & 42.0                       \\
                             & SCL(CVPR'21)\cite{zhang2021self}                       & 36.4                      & 38.6                      & 37.5                      & 35.4                      & 37.0                      & 38.9                      & 40.5                      & 41.5                      & 38.7                      & 39.9                       \\
                             & SAGNN(CVPR'21) \cite{xie2021scale}                     & 36.1                      & \textbf{\textcolor{blue}{41.0}}                      & 38.2                      & 33.5                      & 37.2                      & 40.9                      &\textbf{\textcolor{blue}{ 48.3}}                      & 42.6                      & 38.9                      & 42.7                       \\
                             & NERTNet(CVPR'22)\cite{liu2022learning}                  & \textbf{\textcolor{blue}{38.3}}                      & 40.4                   & \textbf{\textcolor{blue}{39.5}}                      & \textbf{\textcolor{blue}{38.1 }}                     & \textbf{\textcolor{blue}{39.1}}                      & \textbf{\textcolor{blue}{42.3}}                      & 44.4                      & \textbf{\textcolor{blue}{44.2}}                      & 41.7                    & \textbf{\textcolor{blue}{43.2}}                   \\
\multirow{-6}{*}{ResNet-101} & \cellcolor[HTML]{C0C0C0}IPMT(ours) & \cellcolor[HTML]{C0C0C0}\textbf{\textcolor{red}{40.5}} & \cellcolor[HTML]{C0C0C0}\textbf{\textcolor{red}{45.7}} & \cellcolor[HTML]{C0C0C0}\textbf{\textcolor{red}{44.8}} & \cellcolor[HTML]{C0C0C0}\textbf{\textcolor{red}{39.3}} & \cellcolor[HTML]{C0C0C0}\textbf{\textcolor{red}{42.6}} & \cellcolor[HTML]{C0C0C0}\textbf{\textcolor{red}{45.1}} & \cellcolor[HTML]{C0C0C0}\textbf{\textcolor{red}{50.3}} & \cellcolor[HTML]{C0C0C0}\textbf{\textcolor{red}{49.3}} & \cellcolor[HTML]{C0C0C0}\textbf{\textcolor{red}{46.8}} & \cellcolor[HTML]{C0C0C0}\textbf{\textcolor{red}{47.9}} \\ \bottomrule
\end{tabular}}
% \vspace{-2mm}
\end{table*}

\paragraph{Quantitative comparison.}
As shown in Table \ref{Voccompar} and \ref{Cococompar}, we compare our method with previous works on both PASCAL-$5^{i}$ and COCO-$20^{i}$, respectively. It can be found that our IPMT surpasses all other approaches by a large margin and achieves new state-of-the-art results. On PASCAL-$5^{i}$, when using ResNet-50 as the backbone, our proposed IPMT improves the mIoU score by 2.6 under the 1-shot setting compared with the previous best result.
Additionally, we also achieve 1.8 mIoU (under the 1-shot setting) and 2.2 mIoU (5-shot setting) improvements using the ResNet-101 backbone. As for COCO-$20^{i}$, our method with the ResNet-50 backbone outperforms the previous best results by 1.6 and 0.6 mIoU under the two settings, respectively. When using the ResNet-101 backbone, our IPMT improves the mIoU score by 3.5 and 4.7 over the previous best results. These improvements demonstrate the competitiveness of our model on complex data. In addition, we report the comparison with some advanced approaches in terms of the FB-IoU score in Table \ref{fbcompar}, which also shows our superiority.

% \vspace{-3mm}
\paragraph{Quantitative Result.}
We visualize some prediction results of our method and a support-only FSS method (\ie, CyCTR \cite{zhang2021few}) in Figure~\ref{fig:compar}. It can be observed that our results (the 3 row) could effectively relieve the segmentation error caused by the inherent intra-class diversity compared with the results of only using the support information (the 2th row).

\vspace{-1mm}
\subsection{Ablation Study}
\vspace{-1mm}

% A series of extensive ablation studies is conducted to investigate influence of each components for few-shot segmentation performance. 
In this section, we report ablation study results on PASCAL-$5^{i}$ with the ResNet-50 backbone under the 1-shot setting.
% \vspace{-3mm}
\paragraph{Effectiveness of IPM.} To demonstrate the necessity of bridging the category information gap between query and support images using the proposed IPM, we conduct ablation study by learning the category information only from support or query images. Furthermore, we report the performance comparison both with and without iteration. All the ablation studies are conducted with both DSL and QA for a fair comparison.
% to investigate the performance of iteration on aggregating information. 
As shown in Table~\ref{Decoderstr},  `Support only' and `Query only' mean the learnable prototype is only updated by the support feature and the query feature, respectively. `Intermediate' means using both support and query information in our IPM. We observe that our method surpasses other schemes by 1.6 mIoU when not using iteration. This improvement even increases to 3.4 when using iteration. These results clearly demonstrate the effectiveness of our IPM  and also indicate that using iteration is an effective booster for our IPM.
% successfully bridges the category information gap between query and support with deterministic and query-adaptive information. Meanwhile, iteration is a necessity operation to aggregate such information.

\paragraph{Ablation on Different Numbers of IPMT Layers.}
% We can stack more IPMT layers to increase its capacity and validate the effectiveness of our model. 
We vary the number of IPMT layers from 1 to 5 and report the results in Table~\ref{layer}. It shows that using more layers can gradually improve the model performance and using five layers achieves as much as 2.7 mIoU improvement. We did not try more layers considering the accuracy-efficiency trade-off.

% \paragraph{Effect of Query Activation Module.}
% After obtaining the intermediate prototype in \eqref{IPM}, we first attempt to leverage the attention
% scheme to activate query feature map, i.e., treating query feature as Query and intermediate prototype as Key \& Value to make a attention by replacing softmax with sigmoid. However, such a activation scheme does not gain the performance, so we attribute to concatenate the query feature map with expanded intermediate prototype to activate query feature followed by two convolutional layers and yield superior results (see Table~\ref{QAM}).
% \vspace{-3mm}

% \vspace{-3mm}
\paragraph{Effectiveness of DSL and QA.}
% In our proposed IPMT, one layer consists of two steps and a loss, \ie, IPM, QA, and SDL.
% , and each of them plays an important role to aggregate the category information.
Since the effectiveness of the IPM has been proved in Table~\ref{Decoderstr}, here we conduct an ablation study to validate the effectiveness of QA and DSL. We remove all the three components from our model as the baseline model (only with PA), which only uses the support prototype to directly segment the target object like \cite{tian2020prior}. As Table~\ref{diffcom} shows, compared to the baseline, solely using IPM leads to 5.3 mIoU drop. However, when DSL is added, our model achieves 4.1 mIoU improvement over the baseline. This phenomenon is reasonable because there is no guarantee that the learnable prototype in IPM will learn intermediate category knowledge without DSL. Meanwhile, using the QA to activate the query feature map leads to further 2.5 mIoU improvement. These results clearly verify the effectiveness of our proposed QA and DSL.
% \vspace{-3mm}

% \end{wraptable}
% \end{minipage}
% \qquad
% \begin{minipage}{0.45\linewidth}  
% % \begin{wraptable}{r}{6.5cm}
% 	\centering
% 	\caption{Ablation studies that validate the effect of QA.}
% 	\label{QAM}
% 	\resizebox{\linewidth}{!}{
% \begin{tabular}{@{}c|cccc|c@{}}
% \toprule
% Method      & Fold0 & Fold1 & Fold2 & Fold3 & \textbf{Mean} \\ \midrule
% attention   & 67.5  & 72.1  & 58.4  & 56.8  & 63.7          \\
% concatenate & 72.8  & 73.7  & 59.2   & 61.6  & \textbf{66.8}             \\ \bottomrule
% \end{tabular}}
% \end{minipage}
% \begin{wraptable}{r}{6cm}
% \vspace{-3mm}
% 	\centering
% 	\caption{Performance comparison of varying the number of IPMT layers.}
% 	\label{layer}
% 	\resizebox{0.85\linewidth}{!}{
% \begin{tabular}{@{}c|ccccc@{}}
% \toprule
% Layers  & 1 & 2 & 3 & 4 & 5 \\ \midrule
% mIoU   & 64.1 & 64.7 & 65.2 & 65.6 & \textbf{66.8} \\ \bottomrule
% \end{tabular}}
% \vspace{0mm}
% \end{wraptable}
\begin{table}[]
\vspace{-1mm}
\begin{minipage}{0.42\linewidth} 
    % \begin{minipage}{\linewidth}  

% \vspace{0mm}
	\centering
	\caption{Comparison with state-of-the-arts on PASCAL-$5^{i}$ in terms of FB-IoU under the 1-shot and 5-shot settings.}
	\label{fbcompar}
	\resizebox{\linewidth}{!}{
\begin{tabular}{@{}c|c|cc@{}}

\toprule
                             &                                    & \multicolumn{2}{c}{\textbf{FB-IoU}}                   \\
\multirow{-2}{*}{Backbone}   & \multirow{-2}{*}{Methods}          & 1-shot                    & 5-shot                    \\ \midrule
                             & PPNet(ECCV'20)\cite{liu2020part}                     & 69.2                      & 75.7                      \\
                        &PFENet(TPAMI'20)\cite{tian2020prior}                   & 73.3                      & 73.9                      \\
                             & HSNet(ICCV'21)\cite{min2021hypercorrelation}                     & \textbf{\textcolor{blue}{76.7}}                      & \textbf{\textcolor{blue}{80.6}}                      \\
                             & SCL(CVPR'21)\cite{zhang2021self}                       & 71.9                      & 72.8                      \\
                             & DCP(IJCAI'22)\cite{lang2022beyond}                      & 75.6                      & 79.7                      \\
\multirow{-6}{*}{ResNet-50}  & \cellcolor[HTML]{C0C0C0}IPMT(ours) & \cellcolor[HTML]{C0C0C0}\textbf{\textcolor{red}{77.1}} & \cellcolor[HTML]{C0C0C0}\textbf{\textcolor{red}{81.4}} \\ \midrule
                             & A-MCG(AAAI'19)\cite{hu2019attention}                     & 61.2                      & 62.2                      \\
                             & DAN(ECCV'20) \cite{wang2020few}                      & 71.9                      & 72.3                      \\
                   & PFENet(TPAMI'20)\cite{tian2020prior}                   & 72.9                      & 73.5                      \\
                             & CyCTR(NeurIPS'21)\cite{zhang2021few}                  & \textbf{\textcolor{blue}{73.0}}                      & \textbf{\textcolor{blue}{75.4}}                      \\
\multirow{-5}{*}{ResNet-101} & \cellcolor[HTML]{C0C0C0}IPMT(ours) & \cellcolor[HTML]{C0C0C0}\textbf{\textcolor{red}{78.5}} & \cellcolor[HTML]{C0C0C0}\textbf{\textcolor{red}{80.3}} \\ \bottomrule
\end{tabular}}

    \end{minipage}
\qquad
\begin{minipage}{0.52\linewidth}

	\centering
	\caption{Ablation study on the effectiveness of IPM.}
	\label{Decoderstr}
	\resizebox{0.95\linewidth}{!}{
\begin{tabular}{@{}cccc|c@{}}
\toprule
Support only & Query only & Intermediate & Iteration & mIoU \\ \midrule
\Checkmark  &            &       &         & 62.5 \\
             & \Checkmark     &   &          & 59.8    \\
             &     & \Checkmark     &          & 64.1    \\
\Checkmark    &       &         & \Checkmark         & 63.4    \\
      & \Checkmark       &     & \Checkmark         & 60.1    \\
       &        & \Checkmark      & \Checkmark       & \textbf{66.8}    \\ \bottomrule
\end{tabular}}

% \begin{wraptable}{r}{6cm}
\vspace{2mm}
	\centering
	\caption{Performance comparison of varying the number of IPMT layers.}
	\label{layer}
	\resizebox{0.7\linewidth}{!}{
\begin{tabular}{@{}c|ccccc@{}}
\toprule
Layers  & 1 & 2 & 3 & 4 & 5 \\ \midrule
mIoU   & 64.1 & 64.7 & 65.2 & 65.6 & \textbf{66.8} \\ \bottomrule
\end{tabular}}
% \vspace{0mm}
% \end{wraptable}

\end{minipage}
    % \qquad
    % \begin{minipage}{\linewidth}
% \end{wraptable}
% \end{minipage}
% \end{minipage}
% \vspace{-5mm}
\end{table}

\begin{table}[]
\begin{minipage}{0.4\linewidth} 

	\centering
	\caption{Ablation study on the effectiveness of DSL and QA.}
	\label{diffcom}
	\resizebox{0.6\linewidth}{!}{
\begin{tabular}{@{}ccc|c@{}}
\toprule
IPM & DSL & QA & mIoU \\ \midrule
      &         &            & 60.2    \\
\Checkmark       &         &            & 54.9   \\
\Checkmark    & \Checkmark       &            & 64.3    \\
\Checkmark    &  \Checkmark  &   \Checkmark        & \textbf{66.8}    \\ \bottomrule
\end{tabular}}
\end{minipage}
\qquad
\begin{minipage}{0.52\linewidth}

	\centering
	\caption{Intra-class diversity measured by Euclidean distances among the query, support, and intermediate prototypes of four folds on PASCAL-$5^{i}$. 
% 	The results of ‘Mean’ are the averaged distance of all the four folds.
	}
	\label{diverse}
	\resizebox{0.85\linewidth}{!}{
	
\begin{tabular}{c|ccccc}
\toprule
 & fold-0  & fold-1  & fold-2  & fold-3          & Mean   \\ \midrule
$D_{qs}$   & 7.624  & 7.784  & 6.875  & 9.430  & \multicolumn{1}{c}{7.928} \\
$D_{qi}$   & 6.905  & 6.775  & 5.941  & 8.249  & 6.968                     \\
$D_{is}$   & 3.616  & 3.994  & 3.520  & 6.202  & 4.333   \\ \bottomrule

\end{tabular}}
\end{minipage}
% \vspace{2mm}
\end{table}

\begin{figure*}[!hbp]
% \vspace{-1mm}
	\begin{center}
		\includegraphics[scale = 0.15]{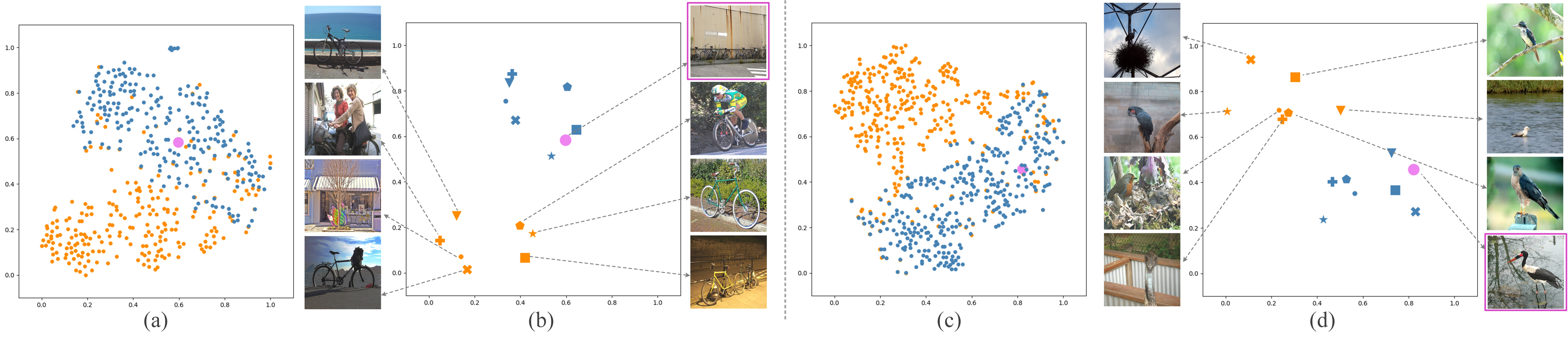}
	\end{center}
	\setlength{\abovecaptionskip}{-2mm}
	\caption{Comparison of the distribution of \textcolor{orange}{support} and \textcolor{blue}{intermediate} prototypes. (a) \& (c): The overall distribution of support prototypes and intermediate prototypes. The latter are closer to the \textcolor{magenta}{query} than the former. (b) \& (d): The visualization of the query images and the support images of some points. Different marks indicate different support prototypes and their corresponding intermediate prototypes.}
	\label{fig:vis}
% 	\vspace{-3mm}
\end{figure*}

% \vspace{-3mm}
\paragraph{Prototype Comparison.}
% To demonstrate that our method effectively relieves the intra-class diversity issue and bridges the category information gap between query and support images,
We first visualize the overall distribution of the support (\textcolor{orange}{orange} points) and intermediate (\textcolor{blue}{blue} points) prototypes given two query (\textcolor{magenta}{magenta} points) images in Figure~\ref{fig:vis} (a) and (c). It is clearly observed that our intermediate prototypes are closer to the query prototypes than the support ones are in the feature space, hence verifying that our method effectively relieves the intra-class diversity issue and bridges the category information gap between query and support images. We also show the support images for some specific points and the query images in Figure~\ref{fig:vis} (b) and (d). Please note that the support and intermediate prototypes of the same support image are shown as two points with the same shape. We find that for these support images which have clear intra-class differences from the query image, our generated intermediate prototypes are successfully pulled to a closer feature space with the query prototype.

Additionally, for evaluating the intra-class diversity objectively, we adopt the Euclidean distance as a metric to measure the distance between the query prototype and the support prototype ($D_{qs}$) in each episode. Then, to further demonstrate the effectiveness of our method, we also measure the distance between the query prototype and the intermediate prototype ($D_{qi}$) and the distance between the intermediate prototype and the support prototype ($D_{is}$). The average distances of each fold and the mean of all the categories on the PASCAL-$5^i$ are shown in Table~\ref{diverse}. From the table, we can clearly see that $D_{qi}$ is smaller than $D_{qs}$ on all folds, which means that our mined intermediate prototype is more similar to the query than the support is. This also demonstrates that our method effectively reduces the distance between the mined prototype and the query and mitigates the intra-class diversity problem.

\vspace{-1.5mm}
\section{Conclusion}
\vspace{-1.5mm}

In this paper, we focus on the intra-class diversity between query and support and introduce an intermediate prototype to bridge the category information gap between them. The core idea is to use the intermediate prototype to aggregate the support-deterministic and query-adaptive category information by our designed Intermediate Prototype Mining Transformer (IPMT) in an iterative way. Surprisingly, despite its simplicity, our method outperforms previous state-of-the-art results by a large margin on two FSS benchmark datasets. We hope our work could inspire future research to concentrate more on the intra-class diversity in FSS.

\section*{Acknowledgments}
This work was supported in part by the National Key R\&D Program of China under Grant 2020AAA0105701; the National Natural Science Foundation of China under Grant 62036011, U20B2065, 32036005, 6202781, 62136007.

{
\small

\bibliographystyle{ieee_fullname}
\bibliography{egbib}

\begin{thebibliography}{10}\itemsep=-1pt

\bibitem{ba2016layer}
Jimmy~Lei Ba, Jamie~Ryan Kiros, and Geoffrey~E Hinton.
\newblock Layer normalization.
\newblock {\em arXiv preprint arXiv:1607.06450}, 2016.

\bibitem{boudiaf2021few}
Malik Boudiaf, Hoel Kervadec, Ziko~Imtiaz Masud, Pablo Piantanida, Ismail
  Ben~Ayed, and Jose Dolz.
\newblock Few-shot segmentation without meta-learning: A good transductive
  inference is all you need?
\newblock In {\em CVPR}, pages 13979--13988, 2021.

\bibitem{carion2020end}
Nicolas Carion, Francisco Massa, Gabriel Synnaeve, Nicolas Usunier, Alexander
  Kirillov, and Sergey Zagoruyko.
\newblock End-to-end object detection with transformers.
\newblock In {\em ECCV}, pages 213--229, 2020.

\bibitem{cheng2021mask2former}
Bowen Cheng, Ishan Misra, Alexander~G Schwing, Alexander Kirillov, and Rohit
  Girdhar.
\newblock Masked-attention mask transformer for universal image segmentation.
\newblock In {\em CVPR}, pages 1290--1299, 2022.

\bibitem{dosovitskiy2020image}
Alexey Dosovitskiy, Lucas Beyer, Alexander Kolesnikov, Dirk Weissenborn,
  Xiaohua Zhai, Thomas Unterthiner, Mostafa Dehghani, Matthias Minderer, Georg
  Heigold, Sylvain Gelly, et~al.
\newblock An image is worth 16x16 words: Transformers for image recognition at
  scale.
\newblock In {\em ICLR}, 2021.

\bibitem{everingham2010pascal}
Mark Everingham, Luc Van~Gool, Christopher~KI Williams, John Winn, and Andrew
  Zisserman.
\newblock The pascal visual object classes (voc) challenge.
\newblock {\em IJCV}, 88(2):303--338, 2010.

\bibitem{guo2021sotr}
Ruohao Guo, Dantong Niu, Liao Qu, and Zhenbo Li.
\newblock Sotr: Segmenting objects with transformers.
\newblock In {\em ICCV}, pages 7157--7166, 2021.

\bibitem{hariharan2011semantic}
Bharath Hariharan, Pablo Arbel{\'a}ez, Lubomir Bourdev, Subhransu Maji, and
  Jitendra Malik.
\newblock Semantic contours from inverse detectors.
\newblock In {\em ICCV}, pages 991--998, 2011.

\bibitem{he2016deep}
Kaiming He, Xiangyu Zhang, Shaoqing Ren, and Jian Sun.
\newblock Deep residual learning for image recognition.
\newblock In {\em CVPR}, pages 770--778, 2016.

\bibitem{hu2019attention}
Tao Hu, Pengwan Yang, Chiliang Zhang, Gang Yu, Yadong Mu, and Cees~GM Snoek.
\newblock Attention-based multi-context guiding for few-shot semantic
  segmentation.
\newblock In {\em AAAI}, volume~33, pages 8441--8448, 2019.

\bibitem{lang2022beyond}
Chunbo Lang, Binfei Tu, Gong Cheng, and Junwei Han.
\newblock Beyond the prototype: Divide-and-conquer proxies for few-shot
  segmentation.
\newblock In {\em IJCAI}, pages 1024--1030, 2022.

\bibitem{li2021adaptive}
Gen Li, Varun Jampani, Laura Sevilla-Lara, Deqing Sun, Jonghyun Kim, and
  Joongkyu Kim.
\newblock Adaptive prototype learning and allocation for few-shot segmentation.
\newblock In {\em CVPR}, pages 8334--8343, 2021.

\bibitem{lin2014microsoft}
Tsung-Yi Lin, Michael Maire, Serge Belongie, James Hays, Pietro Perona, Deva
  Ramanan, Piotr Doll{\'a}r, and C~Lawrence Zitnick.
\newblock Microsoft coco: Common objects in context.
\newblock In {\em ECCV}, pages 740--755, 2014.

\bibitem{liu2021anti}
Binghao Liu, Yao Ding, Jianbin Jiao, Xiangyang Ji, and Qixiang Ye.
\newblock Anti-aliasing semantic reconstruction for few-shot semantic
  segmentation.
\newblock In {\em CVPR}, pages 9747--9756, 2021.

\bibitem{liu2020crnet}
Weide Liu, Chi Zhang, Guosheng Lin, and Fayao Liu.
\newblock Crnet: Cross-reference networks for few-shot segmentation.
\newblock In {\em CVPR}, pages 4165--4173, 2020.

\bibitem{liu2022learning}
Yuanwei Liu, Nian Liu, Qinglong Cao, Xiwen Yao, Junwei Han, and Ling Shao.
\newblock Learning non-target knowledge for few-shot semantic segmentation.
\newblock In {\em CVPR}, pages 11573--11582, 2022.

\bibitem{liu2020part}
Yongfei Liu, Xiangyi Zhang, Songyang Zhang, and Xuming He.
\newblock Part-aware prototype network for few-shot semantic segmentation.
\newblock In {\em ECCV}, pages 142--158, 2020.

\bibitem{liu2021swin}
Ze Liu, Yutong Lin, Yue Cao, Han Hu, Yixuan Wei, Zheng Zhang, Stephen Lin, and
  Baining Guo.
\newblock Swin transformer: Hierarchical vision transformer using shifted
  windows.
\newblock In {\em ICCV}, pages 10012--10022, 2021.

\bibitem{lu2021simpler}
Zhihe Lu, Sen He, Xiatian Zhu, Li Zhang, Yi-Zhe Song, and Tao Xiang.
\newblock Simpler is better: Few-shot semantic segmentation with classifier
  weight transformer.
\newblock In {\em ICCV}, pages 8741--8750, 2021.

\bibitem{milletari2016v}
Fausto Milletari, Nassir Navab, and Seyed-Ahmad Ahmadi.
\newblock V-net: Fully convolutional neural networks for volumetric medical
  image segmentation.
\newblock In {\em 3DV}, pages 565--571, 2016.

\bibitem{min2021hypercorrelation}
Juhong Min, Dahyun Kang, and Minsu Cho.
\newblock Hypercorrelation squeeze for few-shot segmentation.
\newblock In {\em CVPR}, pages 6941--6952, 2021.

\bibitem{nguyen2019feature}
Khoi Nguyen and Sinisa Todorovic.
\newblock Feature weighting and boosting for few-shot segmentation.
\newblock In {\em ICCV}, pages 622--631, 2019.

\bibitem{paszke2019pytorch}
Adam Paszke, Sam Gross, Francisco Massa, Adam Lerer, James Bradbury, Gregory
  Chanan, Trevor Killeen, Zeming Lin, Natalia Gimelshein, Luca Antiga, et~al.
\newblock Pytorch: An imperative style, high-performance deep learning library.
\newblock In {\em NeurIPS}, volume~32, pages 8026--8037, 2019.

\bibitem{russakovsky2015imagenet}
Olga Russakovsky, Jia Deng, Hao Su, Jonathan Krause, Sanjeev Satheesh, Sean Ma,
  Zhiheng Huang, Andrej Karpathy, Aditya Khosla, Michael Bernstein, et~al.
\newblock Imagenet large scale visual recognition challenge.
\newblock {\em IJCV}, 115(3):211--252, 2015.

\bibitem{shaban2017one}
Amirreza Shaban, Shray Bansal, Zhen Liu, Irfan Essa, and Byron Boots.
\newblock One-shot learning for semantic segmentation.
\newblock In {\em BMVC}, 2017.

\bibitem{siam2019amp}
Mennatullah Siam, Boris~N Oreshkin, and Martin Jagersand.
\newblock Amp: Adaptive masked proxies for few-shot segmentation.
\newblock In {\em ICCV}, pages 5249--5258, 2019.

\bibitem{sun2020transtrack}
Peize Sun, Yi Jiang, Rufeng Zhang, Enze Xie, Jinkun Cao, Xinting Hu, Tao Kong,
  Zehuan Yuan, Changhu Wang, and Ping Luo.
\newblock Transtrack: Multiple-object tracking with transformer.
\newblock {\em arXiv preprint arXiv:2012.15460}, 2020.

\bibitem{tian2020prior}
Zhuotao Tian, Hengshuang Zhao, Michelle Shu, Zhicheng Yang, Ruiyu Li, and Jiaya
  Jia.
\newblock Prior guided feature enrichment network for few-shot segmentation.
\newblock {\em IEEE TPAMI}, (01):1--1, 2020.

\bibitem{vaswani2017attention}
Ashish Vaswani, Noam Shazeer, Niki Parmar, Jakob Uszkoreit, Llion Jones,
  Aidan~N Gomez, {\L}ukasz Kaiser, and Illia Polosukhin.
\newblock Attention is all you need.
\newblock In {\em NeurIPS}, volume~30, pages 5998--6008, 2017.

\bibitem{wang2020few}
Haochen Wang, Xudong Zhang, Yutao Hu, Yandan Yang, Xianbin Cao, and Xiantong
  Zhen.
\newblock Few-shot semantic segmentation with democratic attention networks.
\newblock In {\em ECCV}, pages 730--746, 2020.

\bibitem{wang2021max}
Huiyu Wang, Yukun Zhu, Hartwig Adam, Alan Yuille, and Liang-Chieh Chen.
\newblock Max-deeplab: End-to-end panoptic segmentation with mask transformers.
\newblock In {\em CVPR}, pages 5463--5474, 2021.

\bibitem{wang2019panet}
Kaixin Wang, Jun~Hao Liew, Yingtian Zou, Daquan Zhou, and Jiashi Feng.
\newblock Panet: Few-shot image semantic segmentation with prototype alignment.
\newblock In {\em ICCV}, pages 9197--9206, 2019.

\bibitem{wang2021pyramid}
Wenhai Wang, Enze Xie, Xiang Li, Deng-Ping Fan, Kaitao Song, Ding Liang, Tong
  Lu, Ping Luo, and Ling Shao.
\newblock Pyramid vision transformer: A versatile backbone for dense prediction
  without convolutions.
\newblock In {\em ICCV}, pages 568--578, 2021.

\bibitem{wu2021learning}
Zhonghua Wu, Xiangxi Shi, Guosheng Lin, and Jianfei Cai.
\newblock Learning meta-class memory for few-shot semantic segmentation.
\newblock In {\em ICCV}, pages 517--526, 2021.

\bibitem{xie2021scale}
Guo-Sen Xie, Jie Liu, Huan Xiong, and Ling Shao.
\newblock Scale-aware graph neural network for few-shot semantic segmentation.
\newblock In {\em CVPR}, pages 5475--5484, 2021.

\bibitem{yang2020prototype}
Boyu Yang, Chang Liu, Bohao Li, Jianbin Jiao, and Qixiang Ye.
\newblock Prototype mixture models for few-shot semantic segmentation.
\newblock In {\em ECCV}, pages 763--778, 2020.

\bibitem{zhang2021self}
Bingfeng Zhang, Jimin Xiao, and Terry Qin.
\newblock Self-guided and cross-guided learning for few-shot segmentation.
\newblock In {\em CVPR}, pages 8312--8321, 2021.

\bibitem{zhang2019pyramid}
Chi Zhang, Guosheng Lin, Fayao Liu, Jiushuang Guo, Qingyao Wu, and Rui Yao.
\newblock Pyramid graph networks with connection attentions for region-based
  one-shot semantic segmentation.
\newblock In {\em ICCV}, pages 9587--9595, 2019.

\bibitem{zhang2019canet}
Chi Zhang, Guosheng Lin, Fayao Liu, Rui Yao, and Chunhua Shen.
\newblock Canet: Class-agnostic segmentation networks with iterative refinement
  and attentive few-shot learning.
\newblock In {\em CVPR}, pages 5217--5226, 2019.

\bibitem{zhang2021few}
Gengwei Zhang, Guoliang Kang, Yi Yang, and Yunchao Wei.
\newblock Few-shot segmentation via cycle-consistent transformer.
\newblock In {\em NeurIPS}, volume~34, pages 21984--21996, 2021.

\bibitem{zheng2021rethinking}
Sixiao Zheng, Jiachen Lu, Hengshuang Zhao, Xiatian Zhu, Zekun Luo, Yabiao Wang,
  Yanwei Fu, Jianfeng Feng, Tao Xiang, Philip~HS Torr, et~al.
\newblock Rethinking semantic segmentation from a sequence-to-sequence
  perspective with transformers.
\newblock In {\em CVPR}, pages 6881--6890, 2021.

\bibitem{zhu2020deformable}
Xizhou Zhu, Weijie Su, Lewei Lu, Bin Li, Xiaogang Wang, and Jifeng Dai.
\newblock Deformable detr: Deformable transformers for end-to-end object
  detection.
\newblock In {\em ICLR}, 2020.

\end{thebibliography}

}

%%%%%%%%%%%%%%%%%%%%%%%%%%%%%%%%%%%%%%%%%%%%%%%%%%%%%%%%%%%%

\section*{Checklist}

%%% BEGIN INSTRUCTIONS %%%
% The checklist follows the references.  Please
% read the checklist guidelines carefully for information on how to answer these
% questions.  For each question, change the default \answerTODO{} to \answerYes{},
% \answerNo{}, or \answerNA{}.  You are strongly encouraged to include a {\bf
% justification to your answer}, either by referencing the appropriate section of
% your paper or providing a brief inline description.  For example:
% \begin{itemize}
%   \item Did you include the license to the code and datasets? \answerYes{See Section~\ref{gen_inst}.}
%   \item Did you include the license to the code and datasets? \answerNo{The code and the data are proprietary.}
%   \item Did you include the license to the code and datasets? \answerNA{}
% \end{itemize}
% Please do not modify the questions and only use the provided macros for your
% answers.  Note that the Checklist section does not count towards the page
% limit.  In your paper, please delete this instructions block and only keep the
% Checklist section heading above along with the questions/answers below.
% %%% END INSTRUCTIONS %%%

\begin{enumerate}

\item For all authors...
\begin{enumerate}
  \item Do the main claims made in the abstract and introduction accurately reflect the paper's contributions and scope?
    \answerYes{See Section~\ref{intro}.}
  \item Did you describe the limitations of your work?
    \answerYes{See Appendix.}
  \item Did you discuss any potential negative societal impacts of your work?
    \answerYes{See Appendix.}
  \item Have you read the ethics review guidelines and ensured that your paper conforms to them?
    \answerYes{}
\end{enumerate}

\item If you are including theoretical results...
\begin{enumerate}
  \item Did you state the full set of assumptions of all theoretical results?
    \answerNA{}
        \item Did you include complete proofs of all theoretical results?
    \answerNA{}
\end{enumerate}

\item If you ran experiments...
\begin{enumerate}
  \item Did you include the code, data, and instructions needed to reproduce the main experimental results (either in the supplemental material or as a URL)?
    \answerYes{The code will be provided in the supplemental material.}
  \item Did you specify all the training details (e.g., data splits, hyperparameters, how they were chosen)?
    \answerYes{See Section~\ref{data} and \ref{Implementation}.}
        \item Did you report error bars (e.g., with respect to the random seed after running experiments multiple times)?
    \answerNo{We followed previous FSS works and reported the averaged results over 5 independent runs but did not report error bars.}
        \item Did you include the total amount of compute and the type of resources used (e.g., type of GPUs, internal cluster, or cloud provider)?
    \answerYes{See Section~\ref{Implementation}.}
\end{enumerate}

\item If you are using existing assets (e.g., code, data, models) or curating/releasing new assets...
\begin{enumerate}
  \item If your work uses existing assets, did you cite the creators?
    \answerYes{See Section~\ref{data}.}
  \item Did you mention the license of the assets?
    \answerNA{}
  \item Did you include any new assets either in the supplemental material or as a URL?
    \answerNo{}
  \item Did you discuss whether and how consent was obtained from people whose data you're using/curating?
    \answerNo{The data is open source.}
  \item Did you discuss whether the data you are using/curating contains personally identifiable information or offensive content?
    \answerNA{}
\end{enumerate}

\item If you used crowdsourcing or conducted research with human subjects...
\begin{enumerate}
  \item Did you include the full text of instructions given to participants and screenshots, if applicable?
    \answerNA{}
  \item Did you describe any potential participant risks, with links to Institutional Review Board (IRB) approvals, if applicable?
    \answerNA{}
  \item Did you include the estimated hourly wage paid to participants and the total amount spent on participant compensation?
    \answerNA{}
\end{enumerate}

\end{enumerate}

\section*{Appendix}
\appendix

% Optionally include extra information (complete proofs, additional experiments and plots) in the appendix.
% This section will often be part of the supplemental material.
\section{Limitation and Societal Impacts}
\paragraph{Limitation.}
We observe that the performance of our method does not achieve a significant improvement under the 5-shot setting. We argue that this is reasonable because as the number of the support images increases, their internal diversity probably increases. As a result, the intra-class diversity between support and query images probably decreases and
% average prototype will be closer to the overall category prototype, so 
their category information gap will be narrowed. Thus, the ability of our method to correct the support-query deviation will be weakened. However, our work still performs very effectively under the 1-shot setting and provides a new perspective for future research.

\paragraph{Societal Impacts.}
The method proposed in this paper is applied for few-shot semantic segmentation which has many applications in robot vision systems, image recognition software, etc. Compared with previous methods, although our method achieves a large performance improvement, it is still hard to come up with the human cognitive ability. For some safety-critical fields, such as medical treatment and autonomous driving, it is risky if we apply few-shot semantic segmentation methods directly.

% We argue that one potential negative societal impact is that our method may be applied in the military field to facilitate the identification and strike some military targets in the case of strict confidentiality measures and obtaining only a few samples.

% We argue that one potential negative societal impact of our method is the environmental problem which is caused by the training of transformer-like layers. As we all know, compared with the convolution-based networks, transformer-based networks are usually large and computation-expensive with high training resource consumption. Thus, training and deploying such a large model could exacerbate global warming.

\section{Visualization of Extensive Ablative Analysis}
% \subsection{Additional qualitative results.}
\begin{figure*}[thbp]
	\begin{center}
		\includegraphics[scale = 0.73]{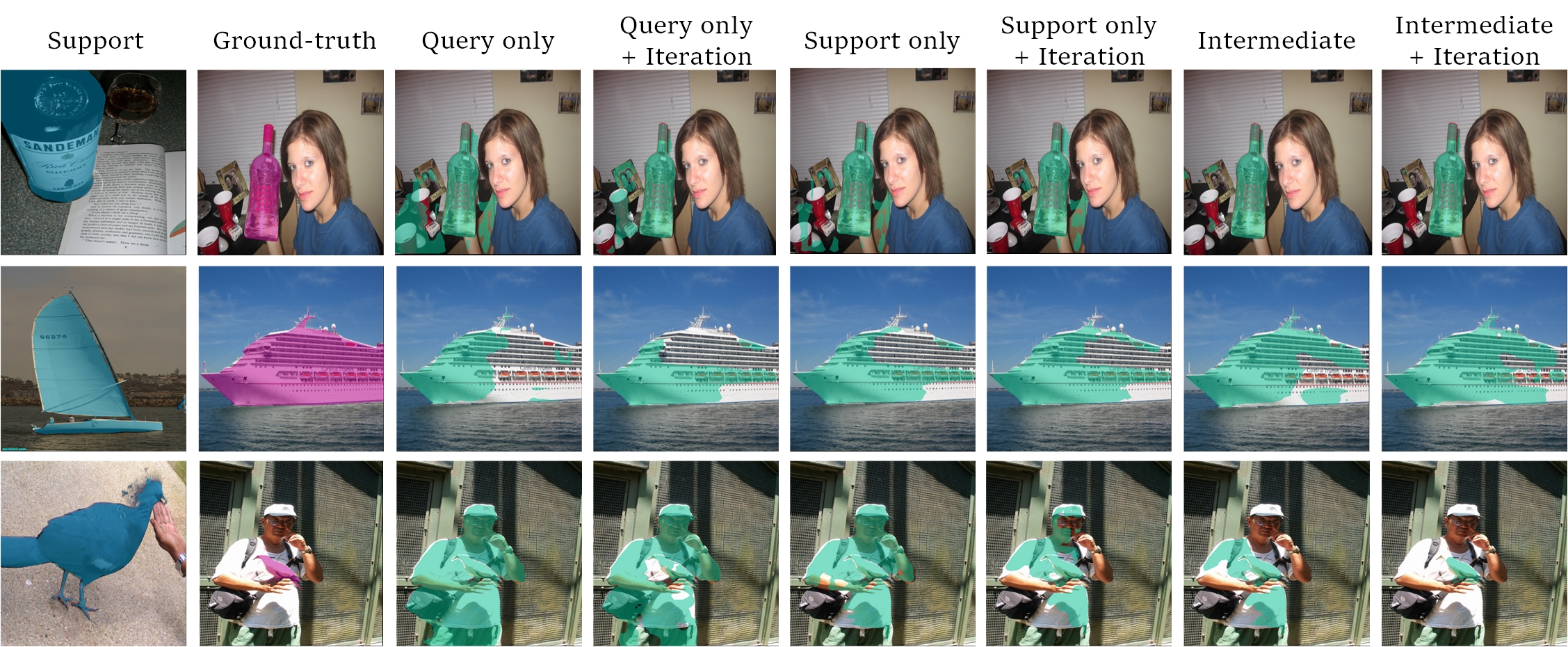}
	\end{center}
	\setlength{\abovecaptionskip}{0.cm}
	\caption{Visualization of extensive ablative analysis on the effectiveness of IPM.}
	\label{fig:ipm}
	\vspace{-3mm}
\end{figure*}

\paragraph{Effectiveness of IPM.}
As a supplement, we show some qualitative comparison results to prove the effectiveness of our IPM in an intuitionistic way. From left to right in Figure \ref{fig:ipm}, the 3$^{rd}$ and 4$^{th}$ columns show the predictions of only using the query feature to update the learnable prototype. We find that the predicted masks are often confused with other distracting objects. Predictions generated by only using the support feature are shown in the 5$^{th}$ and 6$^{th}$ columns, which also can not handle the intra-class difference between support and query images.
Using the intermediate prototype proposed in IPM can relieve the bias of the unidirectional category knowledge. The last column shows that we get the best results when using the intermediate prototype with iterations.

\begin{figure*}[thbp]
	\begin{center}
		\includegraphics[scale = 0.83]{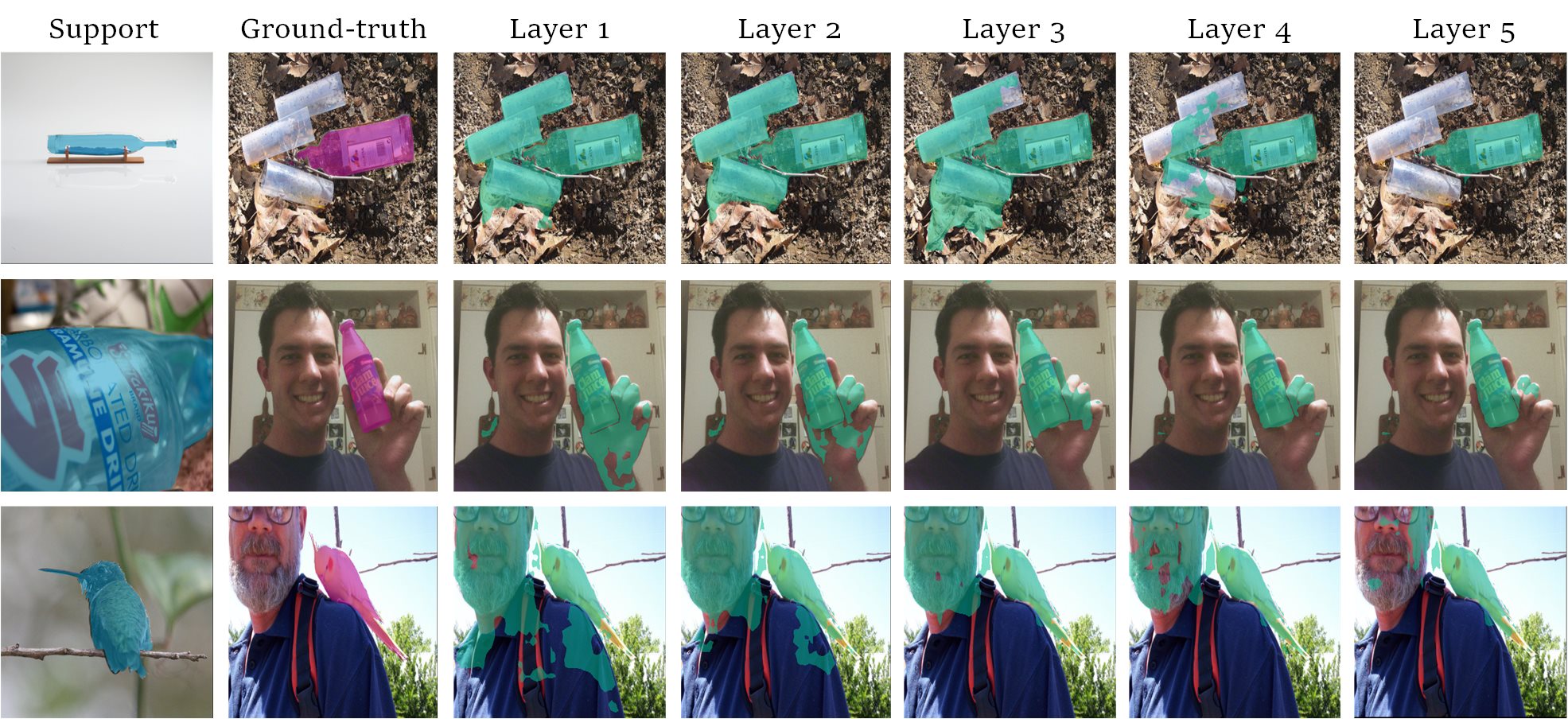}
	\end{center}
	\setlength{\abovecaptionskip}{0.cm}
	\caption{Visualization of the IPMT results by stacking different layers.}
	\label{fig:layers}
	\vspace{-3mm}
\end{figure*}

\paragraph{Visualization of Using Different Numbers of IPMT Layers.}
We further show some qualitative results of stacking different IPMT layers in Figure \ref{fig:layers} to prove the effectiveness of our proposed iterative prototype mining scheme. From the 3$^{rd}$ column to the 7$^{th}$ column, we observe that the accuracy of the predictions is gradually improved by stacking more IPMT layers and the distraction from other objects is relieved. When setting the number of the layers to five, the prediction quality is the best.

\section{Additional Qualitative Results}
\begin{figure*}[thbp]
	\begin{center}
		\includegraphics[scale = 0.73]{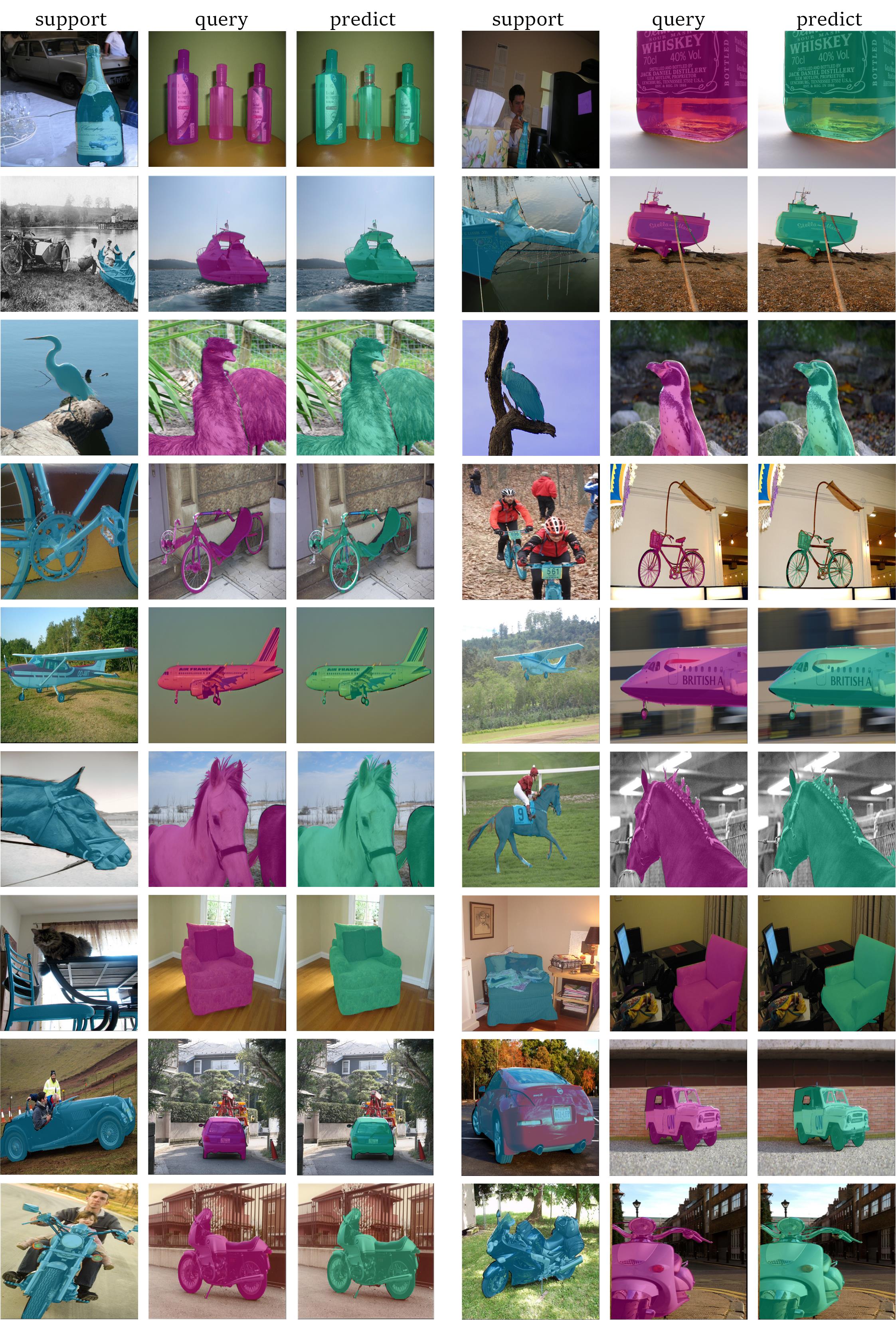}
	\end{center}
	\setlength{\abovecaptionskip}{0.cm}
	\caption{More qualitative results of our IPMT.}
	\label{fig:addcomp}
	\vspace{-3mm}
\end{figure*}

We give more qualitative results in Figure \ref{fig:addcomp} to show the good performance of our IPMT.

\end{document}